\documentclass{article}

\usepackage{arxiv}

\usepackage[utf8]{inputenc} 
\usepackage[T1]{fontenc}    
\usepackage{hyperref}       
\usepackage{url}            
\usepackage{booktabs}       
\usepackage{amsfonts}       
\usepackage{amsmath} 
\usepackage{amssymb}
\usepackage{nicefrac}       
\usepackage{microtype}      
\usepackage{xcolor}         
\usepackage{graphicx}
\usepackage{natbib}
\usepackage{threeparttable}
\usepackage{makecell}
\usepackage{bbm}
\usepackage{caption}
\usepackage{multirow}
\usepackage{colortbl}  

\usepackage[ruled]{algorithm2e}

\title{Towards Generalizable Graph Contrastive Learning:\\ An Information Theory Perspective}

\author{
    {\hspace{1mm}Yige Yuan}\thanks{Data Intelligence System Research Center, Institute of Computing Technology, Chinese Academy of Sciences, Beijing, China. <yuanyige20z, xubingbing, shenhuawei, caoqi, cenketing, zhengwen20s, cxq@ict.ac.cn>}\\
	\And
	{\hspace{1mm}Bingbing Xu$^{*}$}\\
	\And
	{\hspace{1mm}Huawei Shen$^{*}$} \\
	\And
	{\hspace{1mm}Qi Cao$^{*}$} \\
	\And
	{\hspace{1mm}Keting Cen$^{*}$} \\
	\And
	{\hspace{1mm}Wen Zheng$^{*}$} \\
	\And
	{\hspace{1mm}Xueqi Cheng}\thanks{ CAS Key Laboratory of Network Data Science and Technology, Institute of Com- puting Technology, Chinese Academy of Sciences, Beijing, China.}\\
}

\date{}

\renewcommand{\shorttitle}{\textit{arXiv} Template}

\begin{document}
\maketitle

\begin{abstract}
Graph contrastive learning (GCL) emerges as the most representative approach for graph representation learning, which leverages the principle of maximizing mutual information (InfoMax) to learn node representations applied in downstream tasks. To explore better generalization from GCL to downstream tasks, previous methods heuristically define data augmentation or pretext tasks. However, the generalization ability of GCL and its theoretical principle are still less reported. In this paper, we first propose a metric named GCL-GE for GCL generalization ability. Considering the intractability of the metric due to the agnostic downstream task, we theoretically prove a mutual information upper bound for it from an information-theoretic perspective. Guided by the bound, we design a GCL framework named InfoAdv with enhanced generalization ability, which jointly optimizes the generalization metric and InfoMax to strike the right balance between pretext task fitting and the generalization ability on downstream tasks. We empirically validate our theoretical findings on a number of representative benchmarks, and experimental results demonstrate that our model achieves state-of-the-art performance.
\end{abstract}


\section{Introduction} 

Graph representation learning has been widely adopted to improve the performance of numerous scenarios, including traffic prediction \cite{DBLP:conf/ijcai/WuPLJZ19}, recommendation systems \cite{DBLP:conf/sigir/0001DWLZ020}, and drug discovery \cite{DBLP:conf/nips/LiuABG18}. As a powerful strategy for learning low-dimensional embeddings, they can preserve graph attributive and structural features without the need for manual annotation. Among these methods, graph contrastive learning (GCL) \cite{DBLP:conf/nips/SureshLHN21,DBLP:conf/www/0001XYLWW21,DBLP:conf/nips/YouCSCWS20,DBLP:journals/corr/abs-2006-04131,DBLP:conf/iclr/VelickovicFHLBH19} has been the most representative one, which learns node embeddings through maximizing the mutual information between representations of two augmented graph views. Under such principle, namely InfoMax \cite{DBLP:journals/computer/Linsker88}, the learned node representations through GCL have achieved great success when applied in downstream tasks.

To explore better generalization from GCL to downstream tasks, previous methods heuristically or experimentally define data augmentation or pretext tasks. Specifically, \citet{DBLP:conf/www/0001XYLWW21} leverage domain knowledge to remove unimportant structures as augmentations. \citet{DBLP:conf/nips/YouCSCWS20} conduct various experiments and draw conclusions indicating that difficult pretext tasks can improve generalization ability. Similar conclusions have also been explored in computer vision field, \citet{DBLP:journals/corr/abs-2111-06377} find experimentally that harder pretext tasks, i.e., masking an input image with a high percentage, would make the model generalize better. However, these works are still heuristic. As a result, the generalization ability of GCL and its theoretical principle are still less reported. 

In this paper, we propose a metric for GCL generalization ability named GCL-GE inspired by the generalization error of supervised learning \cite{DBLP:journals/corr/abs-2106-10262,DBLP:conf/nips/XuR17,DBLP:conf/aistats/RussoZ16}. However, this metric can hardly be optimized directly due to the agnostic downstream task. To combat this, we theoretically prove a mutual information (MI) upper bound for it from an information-theory perspective. Such upper bound indicates that reducing MI when fitting pretext tasks is beneficial to improve the generalization ability on downstream tasks. In that case, there is a contradiction between our findings and the InfoMax principle of GCL, which reveals that traditional GCL will lead to poor generalization by simply maximizing MI.

Guided by the discovery aforementioned, we design a new GCL framework named InfoAdv with enhanced generalization ability. Specifically, to achieve high generalization ability via minimizing the MI upper bound, we first decompose the bound into two sub-parts based on the Markov data processing inequality \cite{DBLP:journals/tit/Raginsky16}: (1) MI between original graph and its augmentation view; (2) MI between the augmentation view and the representations.  Such decomposition allows the new framework to be flexibly adapted to current GCL 
methods based on data augmentation. We propose a view generator 
and a regularization term based on KL divergence to achieve the goal of two sub-parts respectively. To retain the ability of capturing  the characteristics of graph, our framework is attached to current GCL methods to jointly optimize the generalization metric and InfoMax via an adversarial style, striking the right balance between pretext tasks fitting and generalization ability on downstream tasks.

To demonstrate the effectiveness of our framework, we empirically validate our theoretical findings on a number of representative benchmarks, and experimental results demonstrate that our model achieves state-of-the-art performance. Moreover, extensive ablation studies are conducted to evaluate each component of our framework.

%
%
%

\section{Preliminaries} 

In this section, we briefly introduce the framework of GCL and the generalization error under supervised learning.

\subsection{Graph Contrastive Learning}
An attributed graph is represented as $\mathcal{G}=(\mathcal{V}, \mathcal{E})$ , where $\mathcal{V}=$ $\left\{v_{1}, v_{2}, \cdots, v_{N}\right\}$ is the node set, and $\mathcal{E} \subseteq \mathcal{V} \times \mathcal{V}$ is the edge set. $\boldsymbol{X} \in \mathbb{R}^{N \times D}$ is the feature matrix, where $N$ is the number of nodes and $D$ is the feature dimension. $\boldsymbol{A} \in\{0,1\}^{N \times N}$ is the adjacency matrix, where $\boldsymbol{A}_{i j}=1$ if $\left(v_{i}, v_{j}\right) \in \mathcal{E}$. 

The aim of GCL is obtaining a GNN encoder $f_{\theta}$ which takes $\mathcal{G}$ as input and outputs low-dimensional node embeddings. GCL achieves this goal by mutual information maximization between proper augmentations $t_1$ and $t_2$, namely InfoMax principle \cite{DBLP:journals/computer/Linsker88}. Specifically, taking the pair consisting of the anchor node $v_{i}$ on augmentations $t_1$ and itself $v_{i}^{+}$ on augmentation $t_2$ as a positive pair, and the pair consisting of it and other node $\boldsymbol{v^{-}}=\left\{ v_{j} \mid i \neq j \right\}$ as a negative sample, we need to maximize MI of positive pair $(v_{i}, v_{i}^{+})$, while minimize MI of negative pair $(v_{i},v_{j})$. The learned representations via the trained encoder $f_{\theta}$ is finally used in downstream tasks $\mathcal{T}$ where a new model $q_{\psi}$ is trained under supervision of ground truth label $y_i$ for node $v_i$. We denote the loss function of GCL pretext task and downstream task as $\ell_{un}$ and $\ell_{sup}$. The $f_{\theta}$ and $q_{\psi}$ is optimized as Eq.(\ref{gcl}). In Appendix \ref{Related Work}, we further describe related SOTA GCL methods, and provide detailed comparisons with our method.


\begin{equation}
\begin{split}
f_{\theta^{*}} &= \underset{f_{\theta}}{\arg \min } \sum_{i=1}^{N} \ell_{un} (v_i, v_{i}^{+},\boldsymbol{v^{-}},f_{\theta})\\
q_{\psi^{*}} &=\underset{q_{\psi}}{\arg \min } \sum_{i=1}^{N} \ell_{sup} ( f_{\theta}(\mathcal{G}_{v_{i}}), y_i, q_{\psi}) \label{gcl}
\end{split}
\end{equation}

\subsection{Generalization Error of Supervised Learning}

Define a training dataset $S=\left\{(\boldsymbol{x}_{s}^{i} , {y}_{s}^{i}) \mid i \in N \right\} \in (\mathcal{X} \times \mathcal{Y})^{N}$, which is composed of $N$ i.i.d. samples generated from an unknown data distribution $\mu=P(X, Y)$. $X$ and $Y$ are the random variables of data and labels, and the training dataset $\boldsymbol{X}_S \in \mathbb{R}^{N \times D}, \boldsymbol{Y}_S \in \mathbb{R}^{N}$. Given a learning algorithm $P_{W \mid S}$ taking the training dataset $S$ as input, the output is a hypothesis $w$ of space $\mathcal{W}=\{w: \mathcal{X} \rightarrow \mathcal{Y}\}$ whose random variable is denoted as $W$. The loss function is $\ell: \mathcal{X} \times \mathcal{Y} \times \mathcal{W} \rightarrow \mathbb{R}^{+}$ \cite{DBLP:journals/corr/abs-2106-10262,DBLP:conf/nips/XuR17}.

According to Shwartz and David \cite{DBLP:books/daglib/0033642}, the empirical risk of a hypothesis $\boldsymbol{w} \in \mathcal{W}$ is defined on the training dataset $S$ as $L_{S}(\boldsymbol{w})$ in Eq.(\ref{eq:er_pr_sl1}). The population risk of $\boldsymbol{w}$ is defined on unknown distribution $\mu$ as $L_{\mu}(\boldsymbol{w})$ in Eq.(\ref{eq:er_pr_sl2}). The generalization error is the difference between the population risk and its empirical risk over all hypothesis $W$, which can be defined as $\operatorname{gen}\left(\mu, P_{W \mid S}\right)$ in Eq.(\ref{eq:gen_sl}). In supervised learning, generalization error is a measure of how much the learned model suffers from overfitting, which can be bounded by the input-output MI of the learning algorithm $I(S ; W)$ \cite{DBLP:conf/nips/XuR17}.

	
\begin{gather}
L_{S}(\boldsymbol{w})  \triangleq \frac{1}{N} \sum_{i=1}^{N} \ell\left( \boldsymbol{x}_{s}^{i}, y_{s}^{i},\boldsymbol{w}\right) \label{eq:er_pr_sl1}\\
L_{\mu}(\boldsymbol{w})  \triangleq \underset{(\boldsymbol{x}, y) \sim \mu}{\mathbb{E}} \ell(\boldsymbol{x}, y,\boldsymbol{w}) \label{eq:er_pr_sl2} \\ 
\operatorname{gen}\left(\mu, P_{W \mid S}\right) \triangleq \mathbb{E}\left[L_{\mu}(W)-L_{S}(W)\right] \label{eq:gen_sl}
\end{gather}


\section{MI Upper Bound for GCL-GE}

In this section, we define the generalization error metric for graph contrastive learning (GCL) and prove a mutual information upper bound for it based on information theory.

\subsection{Generalization Error Definition} \label{Error_Definition_Section}

In graph representation learning, the generalization ability is used to measure the model performance difference between pretext task and downstream task, and the model is trained on pretext task. The pretext task is denoted as $\mathcal{P}$ with loss function $\ell_{un}$; The downstream task is denoted as $\mathcal{T}$ with loss function $\ell_{sup}$. Given a GCL algorithm $P_{F \mid \mathcal{P}}$ fitting pretext task $\mathcal{P}$, the output is a hypothesis $f_{\theta}$ of space $\mathcal{F}$ whose random variable is denoted as $F$. We define the GCL's empirical risk of a hypothesis $f_{\theta}$ as $L_{\mathcal{P}}(\mathcal{G},f_{\theta})$ in Eq.(\ref{eq:er}), measuring how precisely does the pretext task fit. The population risk of the same hypothesis is defined as $L_{\mathcal{T}}(\mathcal{G},f_{\theta})$ in Eq.(\ref{eq:pr}), measuring how well the fixed encoder trained by pretext task is adapted to the downstream task. 

\begin{align}
L_{\mathcal{P}}(\mathcal{G},f_{\theta})  &\; \triangleq  \frac{1}{N} \sum_{i=1}^{N} \ell_{un} (v_i, v_{i}^{+},\boldsymbol{v^{-}},f_{\theta}) \label{eq:er}\\
L_{\mathcal{T}}(\mathcal{G},f_{\theta})  &\; \triangleq  \frac{1}{N} \sum_{i=1}^{N} \ell_{sup} ( f_{\theta}(\mathcal{G}_{v_{i}}), y_i, q_{\psi})  \label{eq:pr}
\end{align}

Inspired by the generalization of supervised learning, the GCL generalization error is the difference between the population risk and the empirical risk over all hypothesis $F$. Considering that the two losses may have different scales, an additional constant $s$ related to pretext task is needed to adjust the scale. Thus, the GCL generalization error, using \textbf{\textit{GCL-GE}} as shorthand, are defined as $\operatorname{gen}\left(\mathcal{T}, P_{F \mid \mathcal{P}}\right)$ in Eq.(\ref{eq:ge}).

\begin{equation}
\operatorname{gen}\left(\mathcal{T}, P_{F \mid \mathcal{P}}\right) \; \triangleq \; \mathbb{E}\left[L_{\mathcal{T}}(\mathcal{G},F)-s \cdot L_{\mathcal{P}}(\mathcal{G},F)\right] \label{eq:ge} 
\end{equation}


Overall, the GCL-GE in Eq.(\ref{eq:ge}) is the difference between the population risk and the empirical risk with some scaling transform, which captures the generalization ability of GCL. Recalling that in supervised learning, the goal is to transfer a hypothesis from seen training data to unseen testing data. For GCL, correspondingly, the goal is to transfer a hypothesis from pretext task to downstream task. However, it is worth knowing that the intuitive definition of GCL-GE in Eq.(\ref{eq:ge}) is intractable compared to supervised learning, because of the following two main difficulties. 

(1) Task difficulty: the task of training and testing are the same in supervised learning ($\ell$ in Eq.(\ref{eq:er_pr_sl1}) and Eq.(\ref{eq:er_pr_sl2})), but pretext task and downstream task are different in GCL ($\ell_{un}$ in Eq.(\ref{eq:er}) and $\ell_{sup}$ in Eq.(\ref{eq:pr})) . This problem makes their scales not guaranteed to be the same, making it difficult to measure the difference between population risk and empirical risk. 

(2) Model difficulty: instead of using the same model to train and test in supervised learning, there will be a new model to handle downstream task in GCL, i.e., $q_{\psi}$ in  Eq.(\ref{eq:pr}). The new model fits its own parameters based on the representations to complete the downstream task, making it difficult to evaluate the performance of the representation alone.

\subsection{Generalization Upper Bound} \label{MI_Bound_Section}

In this section, we are devoted to make GCL-GE defined above optimizable, whose first step is to solve the above two difficulties. For this purpose, we introduce the definitions of Latent Class and Mean Classifier, and therefore prove an upper bound of GCL-GE to guide its optimization. The two definitions are as follows: (1) the definition  "Latent Class" follows the contrastive framework proposed by \citet{DBLP:conf/icml/SaunshiPAKK19} and we extend it to graph scenarios. 
Such assumption considers there exists a set of latent classes in GCL pretext task learning and the classes are allowed to overlap and/or be fine-grained. Meanwhile, the classes in downstream tasks are a subset of latent classes. The formal definitions and details are provided in Appendix \ref{2diff}. Such assumption provides a path to formalize how capturing similarity in unlabeled data can lead to quantitative guarantees on downstream tasks. (2) the second definition "Mean Classifier" is leveraged as the downstream model to solve the model difficulty, where the parameters are not learned freely. Instead, they are fixed to be the mean representations of nodes belonging to the same label in downstream tasks, and classifies the nodes in the test set by calculating the distance between each node and each class. With the restrictions on the freedom of downstream model, the model difficulty can be solved. Note that Mean Classifier is a simplification of commonly used downstream models, e.g., Linear regression (LR), and LR has the ability to learn Mean Classifier. The formal definitions and proof of simplification are provided in Appendix \ref{2diff}.

Solving the above two difficulties, We first correlate the pretext task and downstream task, and prove the numerical relationship between their loss (two core elements in GCL-GE) in Lemma 1. Then we prove a mutual information upper bound for GCL-GE  (See Theorem 2) from an information-theoretic point of view. This bound is only related to pretext task and no longer depends on downstream task. Guided by such bound, the GCL-GE can be optimized to improve the generalization performance.




\paragraph{Lemma 1}(Proved in Appendix \ref{Proof_for_Lemma}) Numerical relationship between GCL pretext task loss and downstream task loss. Let $L_{\mathcal{P}}$ denotes the loss function of GCL pretext task, and $L_{\mathcal{T}}^{\mu}$ denotes the loss function of downstream task under Mean Classifier. For any $f_{\theta} \in \mathcal{F}$, if $f_{\theta}(\mathcal{G})$ is $\sigma^{2}$-subgaussian random vector in every direction for every node class $c$, inequality Eq.(\ref{eq:lemma}) holds.

\begin{equation}
L_{\mathcal{P}}(\mathcal{G},f_{\theta}) \leq \gamma(1-\tau) L_{\mathcal{T}}^{\mu}(\mathcal{G},f_{\theta}) + \beta \tau + \epsilon \label{eq:lemma}
\end{equation}
The Eq.(\ref{eq:lemma}) reveals that the scaling parameter $s$ in Eq.(\ref{eq:ge}) is $\gamma(1-\tau)$, which is used to adjust the scale and make the two losses comparable. Noting that, $\gamma=1+c^{\prime} R \sigma \sqrt{\log \frac{R}{\epsilon}}$ and $\beta = 1+c^{\prime} \sigma R$, where $c^{\prime}$ is some constant and $R=\max _{v \in \mathcal{V}}\|f_{\theta}(\mathcal{G}_v)\|$. $\tau$ is the probability of sampling a same latent class twice in the set of all latent classes $\mathcal{C}$ with distribution $\rho$. $\epsilon$ is a noise term.

\begin{equation}
\tau \triangleq \underset{c \pm \sim \rho^{2}}{\mathbb{E}} \mathbf{1}\left[c^{+}=c^{-}\right]=\sum_{c \in \mathcal{C}}[\rho(c)]^{2} .
\end{equation}

\paragraph{Theorem 2}(Proved in Appendix \ref{Proof_for_MIUB}) Mutual Information upper bound for GCL-GE. If the learned representation $F(\mathcal{G})$ is $\sigma^{2}$-subgaussian random vector in every direction for every node class, then the generalization error $\operatorname{gen}\left(\mathcal{T}, P_{F \mid \mathcal{P}}\right)$ is bounded by 
$I(\mathcal{G} ; F)$, MI between the original graph and the GCL learned hypothesis $F$, i.e.
\begin{equation}
 \operatorname{gen}\left(\mathcal{T}, P_{F \mid \mathcal{P}}\right) \leq  \frac{\sqrt{2\sigma^{2}I(\mathcal{G}, F)}-(\beta \tau +\epsilon)}{\gamma(1-\tau)} \label{eq:final}
\end{equation}

\subsection{Contradiction with InfoMax} \label{Contradiction_Section}
The upper bound we proved in section \ref{MI_Bound_Section} provides an information-theoretic understanding of GCL generalization. It reveals that the key to decrease GCL generalization error, i.e., improve generalization performance, is to reduce the MI between original graph and GCL learned hypothesis, i.e., minimize $I(\mathcal{G}, F)$. However, the objective of traditional GCL is to maximize MI between the representation of two views, which can be further proved to have relation with the MI between original graph and GCL learned hypothesis $I(\mathcal{G}, F)$ (Proved in Appendix \ref{Proof_for_Contradiction}). 

Overall, there exists the contradiction between our generalization bound and the traditional GCL's InfoMax principle. In fact, the InfoMax principle of traditional GCL concentrates on pretext task fitting to learn hypothesis, while our bound aims to improve its generalization ability. These two principles involve a trade-off between them, with which there are similar concepts in supervised learning \cite{DBLP:conf/nips/XuR17,DBLP:conf/aistats/RussoZ16}. To improve the performance on downstream task, the key is to capture these two principles of GCL simultaneously. In other word, striking the right balance between pretext task fitting and downstream task generalization by controlling the MI is necessary.

\begin{figure*}[tbp]
  \centering
  \includegraphics[width=0.95\linewidth]{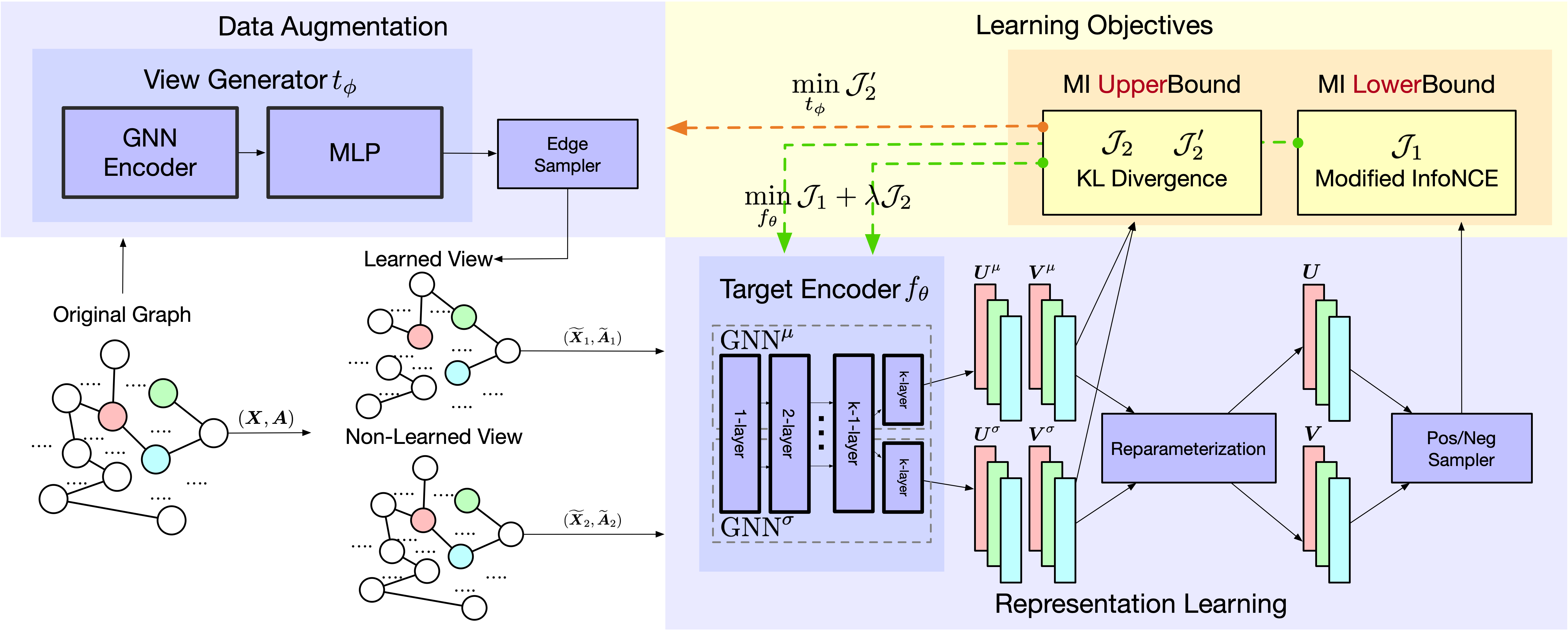}
  \captionsetup{font=small}
  \caption{The key modules of InfoAdv are: (1) a view generator $t_{\phi}$ in data augmentation part, (2) a target encoder $f_{\theta}$ in representation learning part, (3) the learning objectives. $t_{\phi}$ minimizes KL divergence loss to generate augmented views. $f_{\theta}$ minimizes both a KL regularization term and InfoNCE to learn representations. Optimizing $t_{\phi}$ and $f_{\theta}$ jointly can not only capture the characteristics of the graph, but also improve the generalization.} \label{InfoAdv}
\end{figure*}

\section{Methodology: InfoAdv for Generalizable GCL}

In this section, we present our mutual information adversarial framework (\textbf{\textit{InfoAdv}}) in detail, starting with theoretical motivation based on previous sections, followed by the overall framework of InfoAdv, and the last is the instantiation details. 

\subsection{Theoretical Motivation}

InfoAdv is designed for both downstream task generalization and pretext task fitting. Pretext task fitting aims to preserve attribute and structural information in node representations, while downstream task generalization hopes to make the learned representations perform well on downstream tasks. These two goals need to be achieved with the help of both the generalization principle and the InfoMax principle. To make our method can be flexibly adapted to the current GCL framework and thus improve their generalization ability, we implement these two principles based on traditional GCL framework. The generalization principle is achieved from two perspectives: data augmentation and representation learning \cite{DBLP:journals/corr/abs-2102-10757}, and the InfoMax principle is reached by traditional GCL losses. In the following two sections, we discuss them in detail.


 \subsubsection{Generalization Principle}

For the purpose of generalization, we need to reduce the MI between the original graph and the learned hypothesis. Thus the question turns to that: How to decrease $I(\mathcal{G};f_{\theta})$ under the framework of graph contrastive learning? 
The framework of graph contrastive learning can be divided into two parts: (1) generating a new view through data augmentation; (2) learning the representations via a contrastive loss function defined on the original graph and the generated view \cite{DBLP:journals/corr/abs-2102-10757}. Thus we focus on this two core components: the view generator and the contrastive loss function, to reduce MI. We leverage Data Processing Inequality theorem (See Theorem 3) proposed in \cite{DBLP:conf/nips/XuR17,DBLP:journals/tit/Raginsky16} to further achieve our goal:

\paragraph{Theorem 3} For all random variables $X, Y, Z$ satisfying the Markov relation $X \rightarrow Y \rightarrow Z$, the inequality holds as:
\begin{equation}
I(X ; Z) \leq \min \{I(X ; Y), I(Y ; Z)\}
\end{equation}

For GCL, the Markov relation is $\mathcal{G} \rightarrow  \mathcal{G}_{view} \rightarrow f_{\theta}$, the first arrow represents process of data augmentation, and the second arrow represents GNN's representation learning, which is directed by the contrastive loss function. Since $I(\mathcal{G},f_{\theta}) \leq \min \{I(\mathcal{G};\mathcal{G} _{view}),I(\mathcal{G} _{view};f_{\theta})\}$, decreasing $I(\mathcal{G};\mathcal{G} _{view})$ or $I(\mathcal{G} _{view};f_{\theta})$ both can decrease $I(\mathcal{G},f_{\theta})$ to improve GCL generalization performance. To achieve this goal, we propose a view generator to decrease $I(\mathcal{G};\mathcal{G} _{view})$  and a regularization term in the contrastive loss function to decrease $I(\mathcal{G}_{view};f_{\theta})$ simultaneously. These two components bring two perspectives for us to improve generalization. We achieve them by the following way:
 
\begin{itemize}
	\item [(1)] View Generator: a view generator is introduced to minimize $I(\mathcal{G} ;\mathcal{G} _{view})$ by optimizing KL divergence, which is a upper bound \cite{DBLP:conf/icml/PooleOOAT19,DBLP:conf/icml/AlemiPFDS018,DBLP:journals/corr/KipfW16a,agakov2004algorithm}  of MI between original graph and its augmented view. The purpose of this optimization is to find the view that minimizes MI.
	\item [(2)] Loss Regularization: another KL regularization term is introduced to minimize $I(\mathcal{G} _{view};f_{\theta})$. The KL term is an upper bound of MI between input data and its representations, pushing representations to be close to a prior-distribution. The purpose of this optimization is to learn the representation that minimizes MI.
\end{itemize}

 \subsubsection{InfoMax Principle}
For the purpose of InfoMax, we just preserve the traditional objective of GCL, that is, optimizing contrastive loss function termed InfoNCE, which is frequently used for contrastive learning. The InfoNCE has been proved to be a lower bound of MI between the representation of two views \cite{DBLP:conf/iclr/TschannenDRGL20,DBLP:conf/icml/PooleOOAT19,DBLP:journals/corr/abs-1807-03748}.

\subsection{Overall Framework}

The overall framework is shown in Figure \ref{InfoAdv}. There are two learnable components contained in InfoAdv, a \textbf{\textit{View Generator}} $t_{\phi}$ and a contrast-based \textbf{\textit{Target Encoder}} $f_{\theta}$. The view generator only concentrates on the generalization principle, while the target encoder concerns both the generalization and the InfoMax. The objective function can be written as Eq.(\ref{eq:whole_objective}), where $t_{\phi}$ and $f_{\theta}$ play a two-player max-min game. We iteratively optimize these two components by iteratively fixing one to optimize the another. Specifically, if the target encoder is fixed, the objective of view generator is shown as Eq.(\ref{eq:t}). Reversely if the view generator is fixed, the objective of target encoder is shown as Eq.(\ref{eq:f}).

\begin{gather}
V(t_{\phi},f_{\theta}) = \max _{f_{\theta}} \min _{t_{\phi}}  I\left(f_{\theta}(\mathcal{G}) ; f_{\theta}(t_{\phi}(\mathcal{G}))\right)- \lambda \cdot KL(f_{\theta} \| \mathcal{N}(0,1)) \label{eq:whole_objective}\\
t_{\phi}^{*} = \arg \min_{t_{\phi}} I\left(f_{\theta}(\mathcal{G}) ; f_{\theta}(t_{\phi}(\mathcal{G}))\right)\label{eq:t}\\
f_{\theta}^{*} = \arg \max _{f_{\theta}} I\left(f_{\theta}(\mathcal{G}) ; f_{\theta}(t_{\phi}(\mathcal{G}))\right) - \lambda \cdot KL(f_{\theta} \| \mathcal{N}(0,1))\label{eq:f}
\end{gather}

We can see that, the first term of target encoder's objective defined in Eq.(\ref{eq:f}) is the standard InfoMax of GCL, which does not take generalization into consideration. Relying solely on InfoMax will lead to an infinite increase in mutual information. Therefore, a KL regularization term within the target encoder Eq.(\ref{eq:f}) and a view generator Eq.(\ref{eq:t}) are added to prevent MI from increasing unlimitedly to improve the generalization performance. The jointly optimization of InfoMax and generalization form an adversarial framework. Comparing to the traditional GCL optimized only by InfoMax principle and may suffer from poor generalization, InfoAdv can not only capture the characteristics of the graph, but also improve the generalization.

\subsection{Instantiation Details}

This subsection describes the instantiation details of InfoAdv. The algorithm for InfoAdv can be found in Appendix \ref{algor}, Algorithm \ref{InfoAdv_Algorithm}. 
We start with the part of data augmentation as shown in Figure \ref{InfoAdv}, where a view generator is optimized to generate augmented views. Our model does not restrict the type of augmentation, without loss of generality, we choose the most representative edge-dropping as our instantiation. Specifically, The edge-dropping view generator consists of three components: (1) a GNN encoder; (2) an edge score MLP; (3) a probabilistic edge sampler. The GNN encoder firstly takes original graph as input and produces the node embedding matrix $\boldsymbol{H} = \text{GNN} \left(\boldsymbol{X},\boldsymbol{A} \right) $, where node $v_i$ has embedding $\boldsymbol{h}_{i}$. 
Then the embeddings of two nodes connected by an edge are concatenated and sent to MLP to get an edge score  $\boldsymbol{s}_{ij} = \text{MLP}\left(\left[\boldsymbol{h}_{i} ; \boldsymbol{h}_{j} \right] \right)$. 
We normalize the score so that it belongs to the interval $\left[a,b\right]$ representing the edge dropping probability, where $0 \leq a \leq b \leq 1$.
Next, a discrete Bernoulli edge sampler using Gumbel-Max reparameterization trick \cite{DBLP:journals/jmlr/BinghamCJOPKSSH19,DBLP:conf/iclr/MaddisonMT17} is utilized to return an edge-dropping mask $\widetilde{\boldsymbol{R}} \sim \mathcal{B}\left(\boldsymbol{s} \right)$. The mask is then applied to original graph to conduct edge-dropping and get augmented adjacency matrix $\widetilde{\boldsymbol{A}} = \boldsymbol{A} \otimes \widetilde{\boldsymbol{R}}$. Eventually, the view generator generates augmentation views for the following target encoder.

Next to be discussed is the representation learning part, where the target encoder aims to implement both the KL regularization term and the InfoMax. Regularization term aims to push representations to be close to a prior-distribution and we take the uniform Gaussian distribution as the prior distribution. Thus we need to obtain the mean and the variance of representations. As a result, the target encoder $f_{\theta}$ consists of two GNNs, $\text{GNN}^{\mu}$ and $\text{GNN}^{\sigma}$ to learn the mean and the variance respectively. And these two GNNs share all layer parameters except the last layer. We denote the first view of graph as $(\widetilde{\boldsymbol{X}}_1,\widetilde{\boldsymbol{A}}_1)$, the second view as $(\widetilde{\boldsymbol{X}}_2,\widetilde{\boldsymbol{A}}_2)$. Taking the first view whose embeddings denoted as $\boldsymbol{U}$ as example, the view are sent into $f_{\theta}$ to get their mean embeddings $\boldsymbol{U}^{\mu}=\text{GNN}^{\mu}(\widetilde{\boldsymbol{X}}_1,\widetilde{\boldsymbol{A}}_1)$ and variance embeddings $\boldsymbol{U}^{\sigma}=\text{GNN}^{\sigma}(\widetilde{\boldsymbol{X}}_1,\widetilde{\boldsymbol{A}}_1)$. Then reparameterization trick\cite{DBLP:journals/corr/KingmaW13} is utilized for modeling the distributed representation $\boldsymbol{U}=\boldsymbol{U}^{\mu} + \epsilon \; \boldsymbol{U}^{\sigma}, \epsilon \sim \mathcal{N}(0,1)$. The second view whose embeddings denoted as $\boldsymbol{V}$ follows the same procedure.

Finally, as shown in the learning objective part of Figure \ref{InfoAdv}, the loss function of view generator and target encoder are defined in Eq.(\ref{eq:loss_func_f_f}). 
The InfoMax term of target encoder's loss function is adapted from GRACE \cite{DBLP:journals/corr/abs-2006-04131}. The second term of target encoder's loss function $\mathcal{J}_2$ is  the KL term which serves as an upper bound of MI. 

\begin{scriptsize}
\begin{gather}
t_{\phi}^{*} = \arg \min_{t_{\phi}} \mathcal{J}_2^{\prime}, \quad
f_{\theta}^{*} = \arg \min_{f_{\theta}} \mathcal{J}_1+ \lambda \mathcal{J}_2 \label{eq:loss_func_f_f} \\
\mathcal{J}_1  =\frac{1}{2 N} \sum_{i=1}^{N}\left[\ell_{1}\left(\boldsymbol{u}_{i}^{\mu}, \boldsymbol{v}_{i}^{\mu}, \boldsymbol{u}_{i} \right)+\ell_{1}\left(\boldsymbol{v}_{i}^{\mu}, \boldsymbol{u}_{i}^{\mu}, \boldsymbol{v}_{i}\right)\right] \\
\mathcal{J}_2 =\frac{1}{2 N} \sum_{i=1}^{N} \left[\ell_{2}(\boldsymbol{u}^{\mu}_{i},\boldsymbol{u}^{\sigma}_{i})+\ell_{2}(\boldsymbol{v}^{\mu}_{i},\boldsymbol{v}^{\sigma}_{i})\right] \;
\mathcal{J}_2^{\prime} =\frac{1}{N} \sum_{i=1}^{N} \ell_{2}(\boldsymbol{u}^{\mu}_{i},\boldsymbol{u}^{\sigma}_{i})\\
\ell_{1}\left(\boldsymbol{u}_{i}^{\mu}, \boldsymbol{v}_{i}^{\mu}, \boldsymbol{u}_{i}\right)=-\log \frac{e^{\theta\left(\boldsymbol{u}_{i}^{\mu}, \boldsymbol{v}_{i}^{\mu} \right) / \tau}}{e^{\theta\left(\boldsymbol{u}_{i}^{\mu}, \boldsymbol{v}_{i}^{\mu} \right) / \tau}+NEG} \\
NEG=\sum_{k=1}^{N} \mathbbm{1}_{[k \neq i]} e^{\theta\left(\boldsymbol{u}_{i}, \boldsymbol{v}_{k}^{\mu} \right) / \tau}+\sum_{k=1}^{N} \mathbbm{1}_{[k \neq i]} e^{\theta\left(\boldsymbol{u}_{i}^{\mu}, \boldsymbol{u}_{k}^{\mu}\right) / \tau}\\
\ell_{2}(\boldsymbol{u}_{i}^{\sigma},\boldsymbol{u}_{i}^{\mu}) = \frac{1}{2}\left[{(\boldsymbol{u}_{i}^{\sigma})}^{2}+{(\boldsymbol{u}_{i}^{\mu})}^{2}-2\log \left( \boldsymbol{u}_{i}^{\sigma} \right)-1\right]
\end{gather}
\end{scriptsize}


\section{Experiments and Analysis}

This section is devoted to empirically evaluating our proposed instantiation of InfoAdv through answering the following questions. We begin with a brief introduction of the experimental setup, and then we answer the questions according to the experimental results and their analysis.

\begin{itemize}
	\item [(Q1)] Does InfoAdv improve generalization performance?
	\item [(Q2)] How does each generalization component affect InfoAdv's performance?
	\item [(Q3)] How does each key hyperparameter affect InfoAdv?
	\item [(Q4)] How does the update frequency ratio affect InfoAdv?
\end{itemize}

\subsection{Datasets and Setups}

We use two kinds of eight widely-used datasets to study the performance on node classification task and link prediction task, they are: (1) Citation networks including Cora \cite{DBLP:journals/ir/McCallumNRS00}, Citeseer \cite{DBLP:conf/dl/GilesBL98}, Pubmed \cite{DBLP:journals/aim/SenNBGGE08}, Coauthor-CS \cite{DBLP:journals/corr/abs-1811-05868}, Coauthor-Phy \cite{DBLP:journals/corr/abs-1811-05868}. (2) Social networks including WikiCS \cite{DBLP:journals/corr/abs-2007-02901}, Amazon-Photos \cite{DBLP:journals/corr/abs-1811-05868}, Amazon-Computers \cite{DBLP:journals/corr/abs-1811-05868}. Details of datasets and experimental settings can be found in Appendix \ref{datasets} and Appendix \ref{setting}.

\begin{table*}
\centering
\captionsetup{font=small}
  \caption{Summary of performance on node classification in terms of $F_1$-micro score in percentage with standard deviation. The highest performance is highlighted in boldface.}
  \label{Node Classification}
  \scriptsize
  \setlength{\tabcolsep}{3mm}{
  \begin{tabular}{lllllllll}
    \toprule
    Datasets & Cora & Citeseer & Pubmed & WikiCS & A-Photo & A-Com & Co-CS & Co-Phy \\
    \midrule
    RawFeat & 64.23$\pm$1.09 & 64.47$\pm$0.72 & 84.70$\pm$0.25	 & 71.98$\pm$0.00 & 87.37$\pm$0.48 & 78.80$\pm$0.17 & 91.87$\pm$0.11 & 94.44$\pm$0.17      \\
    DeepWalk & 75.16$\pm$0.98 & 52.67$\pm$1.34 & 79.91$\pm$0.33 & 74.35$\pm$0.06 & 89.44$\pm$0.11 & 85.68$\pm$0.06  & 84.61$\pm$0.22 & 91.77$\pm$0.15 \\
    DW+Feat & 77.53$\pm$1.44 & 62.50$\pm$0.61 & 84.85$\pm$0.45 & 77.21$\pm$0.03 & 90.05$\pm$0.08 & 86.28$\pm$0.07 & 87.70$\pm$0.04 & 94.90$\pm$0.09    \\
    \cmidrule(r){1-9}
    GAE & 81.63$\pm$0.72 &	67.35$\pm$0.29	  & 83.72$\pm$0.32 & 70.15$\pm$0.01 & 91.62$\pm$0.13 & 85.27$\pm$0.19 & 90.01$\pm$0.71 & 94.92$\pm$0.07 \\
    VGAE & 82.08$\pm$0.45 &	67.85$\pm$0.65	 & 83.82$\pm$0.14 & 75.63$\pm$0.19 & 92.20$\pm$0.11 & 86.37$\pm$0.21 & 92.11$\pm$0.09 & 94.52$\pm$0.00  \\
    \cmidrule(r){1-9}
    DGI   & 82.60$\pm$0.40	& 68.80$\pm$0.70 &	86.00$\pm$0.10 & 75.35$\pm$0.14 & 91.61$\pm$0.22 & 83.95$\pm$0.47 & 92.15$\pm$0.63 & 94.51$\pm$0.52 \\ 
    GRACE & 83.70$\pm$0.68 & 72.39$\pm$0.21 & 86.35$\pm$0.17 & 79.56$\pm$0.02 & 92.79$\pm$0.11  & 87.20$\pm$0.32 & 93.01$\pm$0.06 & 95.56$\pm$0.05   \\
    GCA & 83.81$\pm$0.91 	& 72.85$\pm$0.17 & 86.61$\pm$0.12  & 80.07$\pm$0.12 & 93.10$\pm$0.07 & 87.70$\pm$ 0.43 & 93.08$\pm$0.06 & 95.61$\pm$0.01    \\
    AD-GCL & 83.93$\pm$0.55	& 72.78$\pm$0.23 & 85.26$\pm$0.41  & 78.06$\pm$0.35 & 91.36$\pm$0.16 & 85.96$\pm$ 0.37 & 93.17$\pm$0.02 & 95.66$\pm$0.06    \\
    \cmidrule(r){1-9}
    \textbf{InfoAdv}  & \textbf{84.82$\pm$0.83} & \textbf{73.00$\pm$0.42} & \textbf{86.87$\pm$0.28} & \textbf{81.41$\pm$0.03} & 93.54$\pm$0.04 &  \textbf{88.27$\pm$0.45} & 93.20$\pm$0.05 & \textbf{95.73$\pm$0.05} \\
    \; \textbf{(w/o) Gen}  & 84.40$\pm$0.99 & 72.51$\pm$0.48 & 86.67$\pm$0.23 & 79.63$\pm$0.48  & 93.26$\pm$0.12  & 88.06$\pm$0.39 & \textbf{93.23$\pm$0.14} & 95.67$\pm$0.09  \\
    \; \textbf{(w/o) Reg}  & 84.36$\pm$0.54 & 72.74$\pm$0.27 & 86.82$\pm$0.25 & 80.97$\pm$0.19 & \textbf{93.80$\pm$0.12} & 87.61$\pm$0.11 & 93.04$\pm$0.06  & 95.65$\pm$0.04  \\
    \cmidrule(r){1-9}
    GCN & 82.69$\pm$1.08	& 72.09$\pm$0.78 & 84.81$\pm$0.40 & 77.19$\pm$0.12 & 92.42$\pm$0.22 & 86.51$\pm$0.54  & 93.03$\pm$0.31 & 95.65$\pm$0.16   \\
    GAT & 83.60$\pm$0.29 & 72.56$\pm$0.44 & 85.27$\pm$0.15 & 77.65$\pm$0.11 & 92.56$\pm$0.35 & 86.93$\pm$0.29  & 92.31$\pm$0.24  & 95.47$\pm$0.15 \\
    \bottomrule
  \end{tabular}}
\end{table*}

\begin{table*}[h]
\captionsetup{font=small}
\centering
  \caption{ Summary of performance on link prediction in terms of Area Under Curve (AUC) and Average Precision (AP) in percentage. The highest performance is highlighted in boldface.}
  \label{Link Prediction}
  \scriptsize
  \setlength{\tabcolsep}{1.8mm}{
\begin{tabular}{llllllllllllllllll}
\toprule
\multirow{2}{*}{Datasets} & \multicolumn{2}{c}{Cora}  & \multicolumn{2}{c}{Citeseer} & \multicolumn{2}{c}{Pubmed} & \multicolumn{2}{c}{WikiCS}   & \multicolumn{2}{c}{A-Photo}& \multicolumn{2}{c}{A-Com} & \multicolumn{2}{c}{Co-CS} & \multicolumn{2}{c}{Co-Phy}  &  \\
\cmidrule(r){2-18}
& AUC  & AP & AUC   & AP & AUC  & AP & AUC  & AP   & AUC  & AP  & AUC & AP & AUC  & AP & AUC  & AP &  \\
\midrule
DGI   & 83.45  & 82.95  & 85.36  & 86.94   & 95.24  & 94.53 & 78.18 & 79.30 & 87.47 & 86.93  & 85.25  & 85.22 & 89.99 & 89.01 & 84.35	& 83.80 &  \\
GRACE & 80.65 & 81.64 & 87.81 & 89.52 & 94.99 & 94.68 & 95.28 & 94.78 & 93.99 & 92.60 & \textbf{94.53} & \textbf{93.87} & 93.82 & 93.44 & 87.95	& 86.50 &  \\
GCA & 83.93   & 83.92 & 88.81  & 89.54   & 95.22  & 94.50  & 94.27 & 94.38  & 96.07 & 95.23 & 93.13  & 93.00   & 94.25  & 93.83 & 85.33 & 81.63 &  \\
AD-GCL  & 80.99 & 82.15 & 89.13 & 89.89 & 96.20 & 95.57 & 94.12  & 93.43  & 85.06  & 82.65  & 86.07   & 85.31   & 93.39 & 93.04   & 88.35 & 87.04 &  \\
\cmidrule(r){1-18}
\textbf{InfoAdv} & \textbf{84.39}  & \textbf{85.96}  & \textbf{93.67}  & \textbf{94.15}  & 96.43 & 96.10 & \textbf{96.71} & \textbf{97.02}  & \textbf{97.25}  & \textbf{96.73} & 93.46  & 92.94  & \textbf{95.02} & \textbf{94.57} & 90.82 & 89.40 &  \\
\; \textbf{(w/o) Gen} & 81.91 & 83.28 & 91.77 & 91.38 & \textbf{96.69} & \textbf{96.28} & 96.45 & 96.66 & 93.67 & 91.98 & 94.17 & 93.49 & 94.80 & 94.24 &  \textbf{92.28} & \textbf{91.33} &\\
\; \textbf{(w/o) Reg} & 82.31 & 83.90 & 82.28 & 85.19 & 95.13 & 94.85 & 95.64 & 95.81 & 95.77 & 94.81 & 93.07 & 92.42 & 93.43 & 92.94 & 88.19 & 86.40 & \\
\bottomrule
\end{tabular}}
\end{table*}

\subsection{Performance on Downstream Tasks (AQ1)}

In this section we empirically evaluate our proposed instantiation of InfoAdv on the following three tasks.

\subsubsection{Generalization Performance on Node Classification Task}
We firstly evaluate InfoAdv on node classification using eight open benchmark datasets against SOTA methods. In each case, InfoAdv learns node representations in a fully self-supervised manner, followed by linear evaluation of these representations. We compare InfoAdv with both unsupervised methods and supervised methods\cite{DBLP:conf/iclr/KipfW17,DBLP:conf/iclr/VelickovicCCRLB18}. The unsupervised methods further include traditional graph embedding methods\cite{DBLP:conf/kdd/PerozziAS14} and SOTA self-supervised learning methods\cite{DBLP:conf/www/0001XYLWW21,DBLP:conf/nips/SureshLHN21,DBLP:journals/corr/abs-2006-04131,DBLP:conf/iclr/VelickovicFHLBH19,DBLP:journals/corr/KipfW16a} (both generative and contrastive). The results is summarized in Table \ref{Node Classification}, we can see that InfoAdv outperforms all methods mentioned above, which verifies the effectiveness of our proposed generalization bound and its principle that a balance should be struck between InfoMax and generalization.

\subsubsection{Generalization Performance on Link Prediction Task}
Considering that our method follow the GCL framework, We then evaluate InfoAdv on link prediction using eight public benchmark datasets against the SOTA graph contrastive methods.  methods. In each case, we split edges in original graph into training set, validation set, and testing set, where InfoAdv is only trained in the training set. We introduce negative edge samples and calculate the link probability via the inner product of the representations. The results are summarized in Table \ref{Link Prediction}, we can see that InfoAdv outperforms all of the baseline methods, also verifying its generalization capability.

\subsubsection{Generalization Performance with Increasing Noise Rate}

We finally conduct generalization analysis for InfoAdv over noisy graph with increasing noise rate. Previous GCL methods fit the pretext task under single InfoMax principle, i.e., to capture as much information as possible from the original graph. However, to improve generalization performance, only the downstream task related information is required, whereas the irrelevant information or the noise information will decrease the generalization performance on the downstream task. Based on this, we verify the generalization ability of GCL model by adding noise, indicating that the generalizable model is more resistant to noise.

\begin{figure*} [h]
  \centering
  \includegraphics[width=0.9\linewidth,height=4cm]{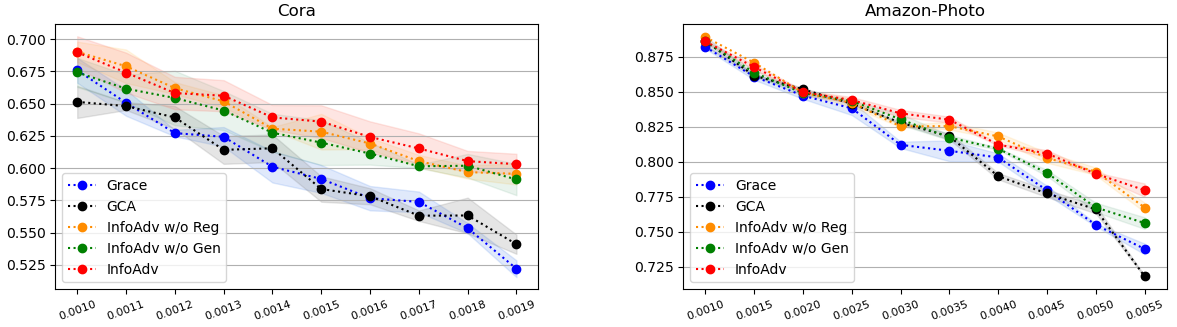}
  \captionsetup{font=footnotesize} 
  \caption{The performance of InfoAdvs and baselines on two datasets with increasing noise rate. The dotted line is the mean scores, the color band is the standard deviation. The x-axis is the hyperparameter for introducing noise, and the noise rate increases synchronously with the hyperparameter. The y-axis is the node classification $F_1$-micro score.} \label{rub}
\end{figure*}

As the results shown in Figure \ref{rub}, we have two key observations: (1) with the rising noise rate, InfoAdv and its variants (w/o Gen and w/o Reg) perform stable, compared to GRACE \cite{DBLP:journals/corr/abs-2006-04131} and GCA\cite{DBLP:conf/www/0001XYLWW21} who drop sharply. (2) Within the different generalization components of  InfoAdv, the performance of InfoAdv outperforms InfoAdv w/o Gen and InfoAdv w/o Reg. The two observation support that, our method shows powerful generalization ability in the noisy graph, which is more challenging for generalization ability.

\subsection{Ablation Studies (AQ2)}

In this section, we study the impact of each generalization component to answer Q2. "(w/o) Gen" denotes without view generator. "(w/o) Reg" denotes without the KL regularization term in target encoder. The results are presented in the bottom of Table \ref{Node Classification} and Table \ref{Link Prediction}, where we can see that both KL term and view generator improve model performance consistently. In addition, the combination of the two generalization components further benefit the performance.

\subsection{Key Hyperparameter Sensitivity (AQ3)}

In this section, we study the sensitivity of two critical hyperparameters: the balance for KL regularization term $\lambda$, and the edge-drop probability of the view learned from view generator $p_{ea}$.
We show the model sensitivity on the transductive node classification task on Cora, under the perturbation of these two hyperparameters, that is, varying $\lambda$ from $1e-07$ to $1e+04$ and $p_{ea}$ from 0.1 to 0.9 to plot the 3D-surface for performance in terms of $F_1$-micro.

\begin{figure*}[!htbp]
  \centering
  \includegraphics[width=0.85\linewidth]{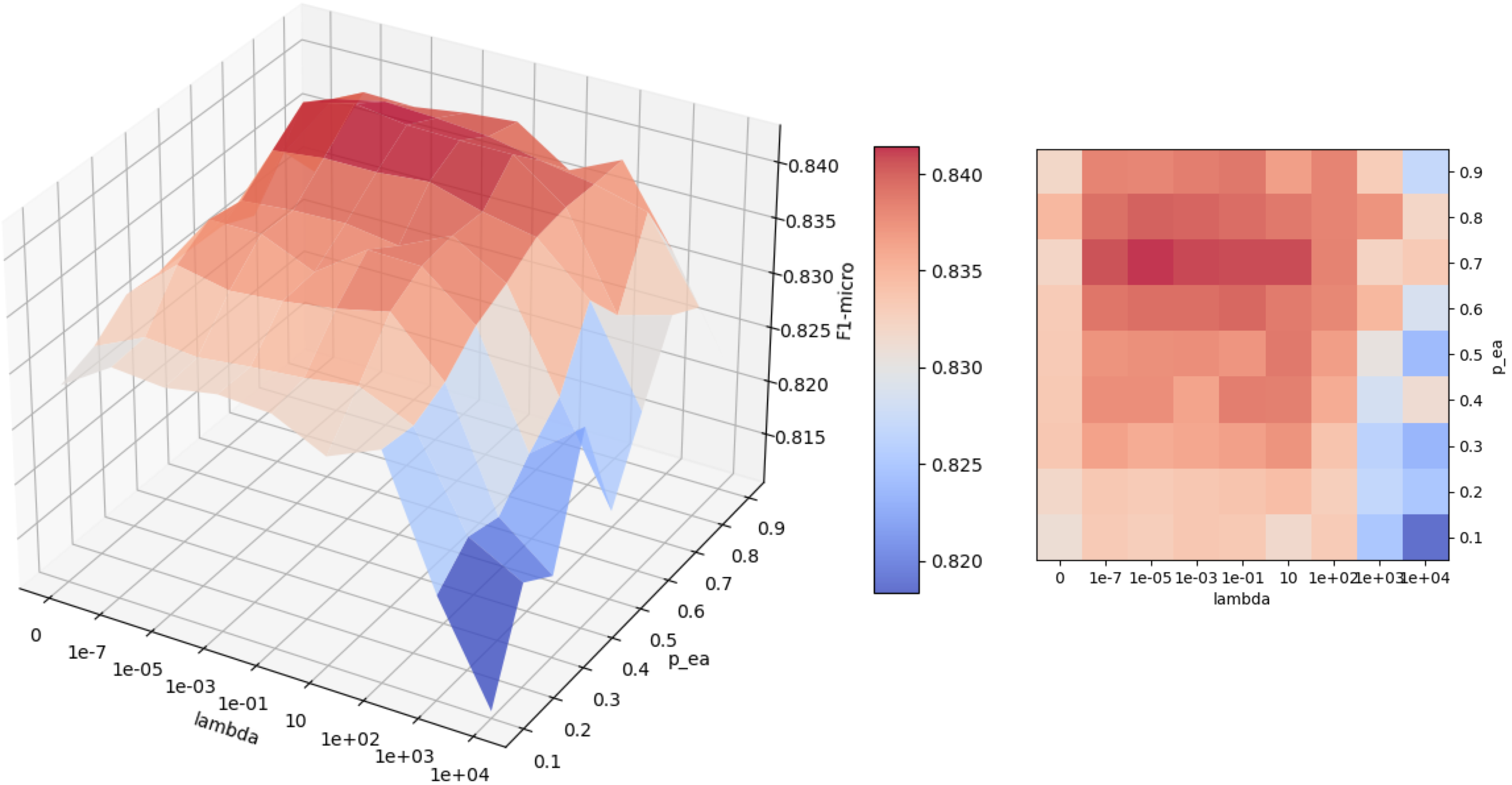}
  \captionsetup{font=small}
  \caption{The \textbf{left} side is the 3D-surface for the performance on Cora dataset, whose three axis are the balance for KL regularization term $\lambda$, the edge-drop probability of the view learned from view generator $p_{ea}$, and the performance $F_1$-micro score. The \textbf{middle} is the colorbar for the performance, indicating that the warmer the color is, the better the model performs. The \textbf{right} side is the heatmap for the 3D-surface when it is observed from the top, i.e., $F_1$-micro axis.} \label{hyper_lamb}

\end{figure*}

The result is shown in Figure \ref{hyper_lamb}, where four key conclusions can be observed,
\begin{itemize}
    \item [(1)] The model is relatively stable when $\lambda$ is not too large, as the model performance remains steady when $\lambda$ varies in the range of $1e-07$ to $1e+02$.
    \item [(2)] The model stays stable even when $\lambda$ is very small, since the model maintains a high level of performance even $\lambda$ reachs as small as $1e-07$.
    \item [(3)] Higher edge-drop probability of the view learned from view generator helps model reach better performance, as the best performance occurs when $p_{ea}=0.7$.
    \item [(4)] The KL regularization component does work. If we do not use the KL term, that is to set $\lambda=0$, the performance drops sharply.
\end{itemize}

\subsection{Update Frequency Ratio (AQ4)}
This section is devoted to analysis update frequency ratio between view generator and decoder. 

\begin{figure*}[h]
  \centering
  \includegraphics[width=0.9\linewidth]{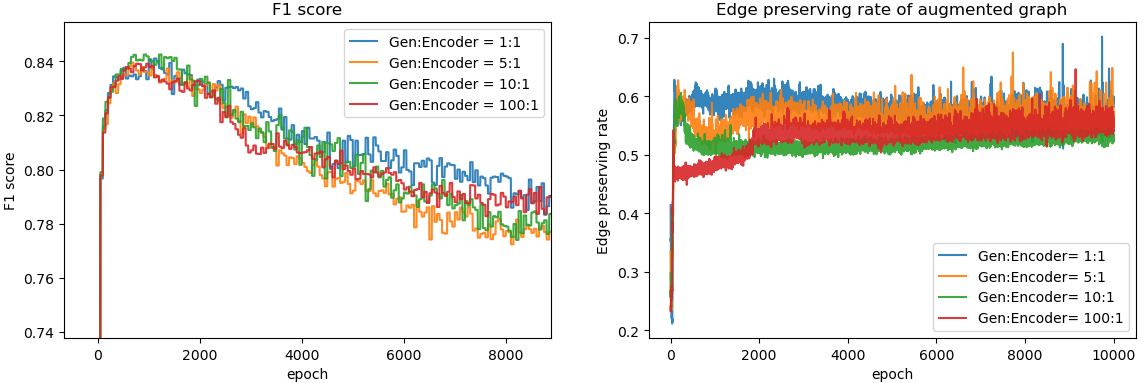}
  \captionsetup{font=footnotesize}
  \caption{The \textbf{left} side is the performance curve of 4 different update frequency ratio, whose x axis is the training epoch and the y axis is the $F_1$ score; The \textbf{right} side is the edge preserving rate curve of the same 4 different update frequency ratio, whose x axis is the training epoch and the y axis is the edge preserving rate} \label{fr}
\end{figure*}

As is shown in the left side of Figure \ref{fr}, we can see that, at the beginning,  with the increasing of the update frequency ratio, i.e., view generator update more and more frequently than decoder, the performance would as well improve and the convergence speed would become faster. when ratio increases up to '10:1', the performance reach the highest. However, when the view generator update too frequently, e.g., up to '100:1' the performance would drop soon.

As is shown in the right side of Figure \ref{fr}, we can see that, with view generator update with more and more high frequency, the edge preserving rate drops as well, indicating that more edges would be removed and making the task more difficult. However, when ratio increases too high up to '100:1', at first, the edge preserving rate dose keep in the lowest level, but with the training continuing, its preserving rate would begin to increase even higher than '10:1'.

These phenomenon may indicates that, the appropriate increase of the update frequency ratio would help to gain better generalization performance. 

\section{Related Work}
In the field of GCL, many work are concentrated on how to do better graph augmentations. Some of them are heuristical, e.g., GCA, GraphCL and JOAO \cite{DBLP:conf/www/0001XYLWW21,DBLP:conf/nips/YouCSCWS20,DBLP:conf/icml/YouCSW21}. Specially, JOAO \cite{DBLP:conf/icml/YouCSW21} dose learn augmentations with an adversarial optimization framework, but their objective is the combination of augmentation types, not the augmentation view. Most recently, AutoGCL \cite{yin2022autogcl} propose a data augmentation strategy. SAIL \cite{yu2022sail} consider the discrepancy of node proximity measured by graph topology and node feature. There are also works focus on the generalization ability of GCL. \cite{trivedi2022understanding} theoretically find that GCL generalization error bound with graph augmentations is linked to the average graph edit distance between classes. AF-GCL \cite{wang2022augmentation} consider GCL generalization for both homophilic graph and heterophilic graph. 

Contrastive information bottleneck \cite{DBLP:conf/nips/SureshLHN21,DBLP:conf/nips/XuCLCZ21,DBLP:conf/nips/Tian0PKSI20}, which is proposed recently dose shares some ideas with our InfoAdv. The common idea of us is the MI minimization. However, there are several fundamental differences as follows.(1) The main contribution of this paper, as well as the original motivation of InfoAdv framework, is the generalization error metric of GCL and its MI upper bound, which, neither AD-GCL \cite{DBLP:conf/nips/SureshLHN21}, InfoGCL \cite{DBLP:conf/nips/XuCLCZ21} nor \citet{DBLP:conf/nips/Tian0PKSI20} involve. (2) As for self-supervised version of Tian et al. and AD-GCL, (despite of the field differences between CV and Graph) they focus only on the MI between two views for contrastive learning, while we concern the MI between GCL input-output. Our method can theoretically cover theirs due to the data processing inequality. (3) Specifically for AD-GCL, they leverage an MI lower bound to conduct MI minimization. However in terms of MI minimization, a single MI lower bound can not guarantee MI is truly minimized. Our method consider MI upper bound, which is more reasonable and could ensure the right MI minimization. (4) As for semi-supervised version of Tian et al. and InfoGCL, they are in need of knowledge from downstream task, which can hardly be satisfied in self-supervised learning. (5) AD-GCL, InfoGCL and Tian et al. define their MI between contrastive views, whereas we define MI between input data and output hypothesis, which is not bound with contrastive learning and can be easily extended to other learning paradigm such as predictive(generative) learning.\cite{DBLP:journals/corr/abs-2102-10757,DBLP:journals/corr/abs-2006-08218}.

Specificcally, for AD-GCL, we provide a more detailed comparison: Theoretically, (1) We provide a metric termed GCL-GE to bridge the generalization gap between GCL upstream and downstream , while AD-GCL does not connect this gap. (2) We prove a realizable bound for GCL-GE optimization even for agnostic downstream task, while AD-GCL does not consider. Methodologically, (1)AD-GCL leverages an MI lower bound (i.e., InfoNCE) to conduct MI minimization. However in terms of MI minimization, a single MI lower bound can not guarantee MI is truely minimized. On the contrary, our method consider MI upper bound (i.e., KL regularization ), which is more reasonable and could ensure the right MI minimization. (2) AD-GCL only improve GCL via data augmentation, while our method consider two components, both data augmentation and representation learning. Experimentally, We modify AD-GCL to node-level classification task and the results show that our InfoAdv outperforms AD-GCL on all eight popular datasets as below. (See Section 5, Table \ref{Node Classification} and Table \ref{Link Prediction})

There are plenty of work done for the generalization error of supervised learning, among which, traditional works draw bounds according to VC dimension, Rademacher complexity \cite{DBLP:books/daglib/0033642} and PAC-Bayesian \cite{DBLP:journals/ml/McAllester99}. Recently, the MI bounds based on information theory \cite{DBLP:conf/aistats/RussoZ16,DBLP:conf/nips/XuR17,DBLP:conf/nips/AsadiAV18,DBLP:conf/colt/SteinkeZ20,DBLP:journals/corr/abs-2106-10262} for generalization error has aroused widespread interests. For the generalization error of contrastive learning, Arora et al. \cite{DBLP:conf/icml/SaunshiPAKK19} propose a learning framework and come up with a Rademacher complexity bounds for its generalization error. Moreover, Nozawa et al. \cite{DBLP:conf/uai/NozawaGG20} adopt PAC bounds based on Arora's framework. Our different with those work are as below: (1) We prove a MI upper bound, which is a better bound compared to Arora et al. and Nozawa et al. (2) We propose a generalization error metric explicitly inspired by supervised learning, which is not mentioned in Arora et al. and Nozawa et al. (3) We extend Arora's framework to the graph(GNN) scenarios.

\section{Conclusion} 

In this paper, we first propose GCL-GE, a metric for GCL generalization ability. Considering the intractability of the metric itself due to the agnostic downstream task, we theoretically prove a mutual information upper bound for it from an information-theoretic perspective. Then, guided by the bound, we design a GCL framework named InfoAdv with enhanced generalization ability, which jointly optimizes the generalization metric and InfoMax to strike the right balance between pretext tasks fitting and the generalization ability on downstream tasks. Finally, We empirically validate our theoretical findings on a number of representative benchmarks, and experimental results demonstrate that our model achieves state-of-the-art performance. Our InfoAdv principle is suitable not only for contrastive learning, but also other learning paradigm like predictive(generative) learning, which will be left to the further exploration.

\newpage
\bibliographystyle{unsrtnat}
\bibliography{main}

\newpage
\appendix

\section{Appendix Summary}
The appendix contains following sections,
\begin{itemize}
    \item [(1)] Proofs: We provide the assumptions for overcoming  difficulties of task and model (Sec. \ref{2diff}); the detailed proof of Lemma 1 (Sec. \ref{Proof_for_Lemma}); the detailed proof of Theorem 2 (Sec. \ref{Proof_for_MIUB}); the detailed proof of contradiction (Sec. \ref{Proof_for_Contradiction});  the detailed proof of loss functions and MI bounds (Sec. \ref{Proof_for_Loss_MI});
    \item [(2)] Descriptions: the summary of datasets (Sec. \ref{datasets}); the detailed experimental settings (Sec. \ref{setting}), including evaluation protocol (subSec. \ref{Eva_Subsection}), baselines (subSec. \ref{Baseline_Subsection}), hyperparameters (subSec. \ref{Hyper_Subsection}) and computing resources (subSec. \ref{Compres_Subsection});
    \item [(3)] Experiemnts: We conduct additional experiments for hyperparameter sensitivity (Sec. \ref{Sensitivity_Section}); additional experiments for GCL-GE metric (Sec.\ref{Metric_Section}); 
    \item [(4)] Others: We discuss the limitations and future explorations (Sec.\ref{Limit_Section}).
\end{itemize}

\section{Overcome Difficulties of Task and Model} \label{2diff}

\subsection{Latent Class}

\paragraph{GCL Pretext Task}
GCL is a typical unsupervised learning method which has no information of downstream real classes. However, it can be assumed that there exists a set of latent classes $\mathcal{C}$  whose probabilities are described by distribution $\rho$. Moreover, a node distribution $\mathcal{D}_c$ is defined over the input graph space $\mathcal{G}$ within each latent class $c \in \mathcal{C}$. GCL samples are generated according to the following scheme,
\begin{itemize}
	\item [(1)] Sample two latent classes $c^{+} c^{-} \sim \rho^{2}$.
	\item [(2)] Sample positive node pair $ (v, v^{+}) \sim (\mathcal{D}_{c^{+}})^{2}$.
	\item [(3)] Sample negative node $ v^{-} \sim \mathcal{D}_{c^{-}}$.
\end{itemize}

In most cases of GCL, the positive pair are comprised of two augmentations of the same node $v_i$, while the negative samples are other nodes $v_{j \; j\neq i}$. We claim that it can be viewed as a set of more fine-grained latent semantic classes.

\paragraph{Downstream Task}
As for the true label from downstream, it can always be a subset of previous latent classes. We will consider the standard supervised learning task of classifying a data point into one of the classes in $\mathcal{C}$. Formally, downstream samples are generated according to the following scheme,

\begin{itemize}
	\item [(1)] Sample one class $c \sim \rho$.
	\item [(2)] Sample node with label $ (v, c) \sim \mathcal{D}_{c}$.
\end{itemize}

Therefore, by optimizing an InfoMax objective which distinguishes these latent classes from each other, the downstream can also be considered. Therefore, the latent class assumption unifies the task from GCL and downstream, which overcomes the task difficulty.

\subsection{Mean Classifier}
The Mean Classifier $q_{\psi}^{\mu}$ is a classifier, whose $k^{\text {th }}$ row parameter is $\boldsymbol{\mu}_{k} \triangleq \underset{v \sim \mathcal{D}_{c_k}}{\mathbb{E}}[f_{\theta}(\mathcal{G}_v)]$, namely the mean $\boldsymbol{\mu}_{c_k}$ of representations of nodes with label $c_k$, for GNN encoder $f_{\theta}$ from GCL and downstream classification task $\mathcal{T}=\left(c_{1}, \ldots, c_{k}\right)$, where node $v_i$ belongs to ground truth label $y_i$.

\begin{equation} 
L_{\mathcal{T}}^{\mu}(\mathcal{G}, f_{\theta}) \triangleq \sum_{i=1}^{N} \ell_{sup} ( f_{\theta}(\mathcal{G}_{v_{i}}), y_i, q_{\psi}^{\mu})
\end{equation}

For GCL, the Mean Classifier can be view as a node clustering, where the parameters of downstream new model are not learned freely. Instead, they are fixed to be the mean representations of nodes with same true label from downstream task. With the restrictions on the freedom of downstream new model, the model difficulty can be solved.

\subsection{Proof for Relation between Mean Classifier and Logistic Regression}
Given a simple scenario of $N$ representation of data $\left\{\boldsymbol{h}_i \mid i \in N\right\}$, with $d$ dimension and $k$ classes $\left\{ c_1, c_2, \cdots, c_k \right\}$. According to the definition of Mean Classifier, the parameter of class $c_k$ is:
\begin{equation}
\boldsymbol{\mu}_{k}=\frac{\sum_{i=1}^{n} \mathbf{1} \left[y=k\right] \cdot \boldsymbol{h}_i}{\sum_{i=1}^{n} \mathbf{1} \left[y=k\right]}
\end{equation}

Given a Logistic Regression classifier, whose weight is shown below.
\begin{equation}
\boldsymbol{W} =\left[\begin{array}{llll}
w_{11} & w_{12} & \cdots & w_{1k}\\
w_{21} & w_{22} & \cdots & w_{2k}\\
\cdots & \cdots & \cdots &\vdots \\
w_{d1} & w_{42} & \cdots & w_{dk}
\end{array}\right]
\end{equation}

The Logistic Regression classifier is exactly a Mean Classifier if we initialize it as Eq.(\ref{mc_init}). The losses of the initial parameter (i.e. the Mean Classifier) is denoted as $L_{\mathcal{T}}^{\mu}$.
\begin{equation}
\boldsymbol{W}_{[:,k]} = \boldsymbol{\mu}_{k} \label{mc_init}
\end{equation}

After initialization, the Logistic Regression classifier will further be optimized according to gradient descent, making the loss lesser than its initial parameter. The optimized losses is denoted as  $L_{\mathcal{T}}$. Finally, the inequality between Mean Classifier and Logistic Regression is shown in Eq.(\ref{mc_lr}).
\begin{equation}
L_{\mathcal{T}} \leq L_{\mathcal{T}}^{\mu} \label{mc_lr}
\end{equation}

\section{Proof for Lemma 1} \label{Proof_for_Lemma}

Our proof follows the framework and important conclusions Eq.(\ref{eq:l_un}), Eq.(\ref{eq:l_un_neq}) and Eq.(\ref{eq:l_un_eq}) from \citet{DBLP:conf/icml/SaunshiPAKK19}. A contrastive loss $L_{\mathcal{P}}(\mathcal{G},f_{\theta})$ can be decomposed into two parts $L_{\mathcal{P}}^{\neq}(\mathcal{G},f_{\theta})$ and $L_{\mathcal{P}}^{=}(\mathcal{G},f_{\theta})$. The first part is the loss suffered when the similar pair and the negative sample come from different classes, describing how well $f_{\theta}$ can distinguish each class. While the second part is when they come from the same class, describing the intraclass deviation of $f_{\theta}$. And for any $f_{\theta} \in \mathcal{F}$, it holds that,
\begin{equation}
L_{\mathcal{P}}(\mathcal{G},f_{\theta})=(1-\tau) L_{\mathcal{P}}^{\neq}(\mathcal{G},f_{\theta})+\tau L_{\mathcal{P}}^{=}(\mathcal{G},f_{\theta}) \label{eq:l_un}
\end{equation}

For $f_{\theta} \in \mathcal{F}$, if the random variable $f_{\theta}(\mathcal{G})$ is $\sigma^{2}$-subgaussian random vector in every direction for every node class $c$ and has maximum norm $R=\max _{v \in \mathcal{V}}\|f_{\theta}(\mathcal{G}_v)\|$. Let $\gamma=1+c^{\prime} R \sigma \sqrt{\log \frac{R}{\epsilon}}$ where $c^{\prime}$ is some constant, and $L_{\gamma,  \mathcal{T}}^{\mu}\left(\mathcal{G}, f_{\theta}\right)$ is $L_{\mathcal{T}}^{\mu}\left(\mathcal{G}, f_{\theta}\right)$ when $\ell_{\gamma}(x)=(1-x / \gamma)_{+}$ is the loss function. Then for all $\epsilon>0$,

\begin{equation}
L_{\mathcal{P}}^{\neq}(\mathcal{G},f_{\theta}) \leq \gamma L_{\gamma, \mathcal{T}}^{\mu}(f_{\theta})+\epsilon \label{eq:l_un_neq}
\end{equation}

For all $f_{\theta} \in \mathcal{F}$, Let $s(f_{\theta})$ be the notion of intraclass deviation defined as ($\ref{eq:s(f)}$), where $\sum(f_{\theta}, c)$ is the covariance matrix of $f_{\theta}(\mathcal{G}_v)$ when $v \sim \mathcal{D}_c$. Then (\ref{eq:l_un_eq}) holds where $c^{\prime}$ is a positive constant.

\begin{gather}
s(f_{\theta}) \triangleq \underset{c \sim \nu}{\mathbb{E}}\left[\sqrt{\|\Sigma(f_{\theta}, c)\|_{2}} \underset{v \sim \mathcal{D}_{c}}{\mathbb{E}}\|f_{\theta}(\mathcal{G}_v)\|\right] \label{eq:s(f)}\\
L_{\mathcal{P}}^{=}(\mathcal{G},f_{\theta})-1 \leq c^{\prime} s(f_{\theta}) \label{eq:l_un_eq}
\end{gather}

Here comes our proof for the connection between pretext task and downstream task. If the random variable $f_{\theta}(\mathcal{G})$ is $\sigma^{2}$-subgaussian random vector in every direction for every node class $c$ and has maximum norm $R=\max _{v \in \mathcal{V}}\|f_{\theta}(\mathcal{G})_v)\|$. Then from Eq.(\ref{eq:l_un_eq}) and the properties of subgaussian vector, we can know that for some constant $c^{\prime}$,

\begin{equation}
L_{\mathcal{P}}^{=}(\mathcal{G},f_{\theta}) \leq c^{\prime} \sigma R+1 \label{eq:l_un_eq_2}
\end{equation}


Then together with Eq.(\ref{eq:l_un}), Eq.(\ref{eq:l_un_neq}) and Eq.(\ref{eq:l_un_eq_2}), we can get the inequality below where $\gamma=1+c^{\prime} R \sigma \sqrt{\log \frac{R}{\epsilon}}$ and $\beta = 1+c^{\prime} \sigma R$, and the Lemma 1 is proved.

\begin{equation}
L_{\mathcal{P}}(\mathcal{G},f_{\theta}) \leq \gamma(1-\tau) L_{\mathcal{T}}^{\mu}(\mathcal{G},f_{\theta}) + \beta \tau + \epsilon \label{eq:l_un_<}
\end{equation}

\section{Proof for Theorem 2: GCL-GE MI Upper Bound \label{Proof_for_MIUB}}

Following \citet{DBLP:conf/aistats/RussoZ16} as well as \citet{DBLP:conf/nips/XuR17}, we start with the Donsker-Varadhan variational representation of the relative entropy \cite{DBLP:books/daglib/0035704}. It says, for any two probability measures $\pi, \rho$ on a common measurable space $(\Omega, \mathcal{K})$, inequality Eq.(\ref{eq:dv}) holds, where $D(\pi \| \rho)$ stands for the KL divergence between two distributions $\pi$ and $\rho$. And the supremum is over all measurable functions $K: \Omega \rightarrow \mathbb{R}$, such that $e^{K} \in L^{1}(\rho)$.

\begin{equation}
D(\pi \| \rho)=\sup _{K}\left\{\int_{\Omega} K \mathrm{~d} \pi-\log \int_{\Omega} e^{K} \mathrm{~d} \rho\right\} \label{eq:dv}
\end{equation}

The relation between the expectation of $K(X)$ and its Lebesgue integral is defined below, from where $\mathbb{P}$ is the probability density function of $X$, 
\begin{equation}
\mathbb{E}[K(X)]=\int_{\Omega} K d \mathbb{P} \label{eq:Lebint}
\end{equation}

Let $\pi = P_{\mathcal{G}, F}$, \; $\rho = P_{\mathcal{G}} \otimes P_{F}$, \; $K=L_{\mathcal{P}}$,  we have these three inequalities hold for any $\lambda \in \mathbb{R}$. The step $a$ is from  Eq.(\ref{eq:dv}) and Eq.(\ref{eq:Lebint}). The step $b$ is because of the $\sigma^{2}$-subgaussian assumption of random vector $F(\mathcal{G})$. By the definition of $\sigma^{2}$-subgaussian random vector\footnote{A random vector $V \in \mathbb{R}^{d}$ is $\sigma^{2}$-subgaussian in every direction, if $\forall u \in \mathbb{R}^{d},\|u\|=1$, the random variable $\langle u, V\rangle$ is $\sigma^{2}$-subgaussian.}, we can know that $L_{\mathcal{P}}(\bar{\mathcal{G}}, \bar{F})$ is also $\sigma^{2}$-subgaussian random variable, then by the definition of $\sigma^{2}$-subgaussian variable\footnote{A random variable $\mathrm{X}$ is called $\sigma^{2}$-subgaussian if $\mathbb{E}\left[e^{\lambda(X-\mathbb{E}[X])}\right] \leq e^{\lambda^{2} \sigma^{2} / 2}, \forall \lambda \in \mathbb{R}.$}, we can reach the step $b$.

\begin{equation}
\begin{split}
	 & D\left(P_{\mathcal{G}, F} \| P_{\mathcal{G}} \otimes P_{F}\right) \overset{a}{\geq} \mathbb{E}[\lambda L_{\mathcal{P}}(\mathcal{G}, F)]-\log \mathbb{E}\left[e^{\lambda L_{\mathcal{P}}(\bar{\mathcal{G}}, \bar{F})}\right] \\
	& \overset{b}{\geq} \lambda(\mathbb{E}[L_{\mathcal{P}}(\mathcal{G},F)]-\mathbb{E}[L_{\mathcal{P}}(\bar{\mathcal{G}},\bar{F})])-\frac{\lambda^{2}\sigma^{2}}{2}\\
	& \overset{c}{\geq} \lambda(\mathbb{E}[L_{\mathcal{P}}(\mathcal{G},F)]-\gamma(1-\tau) \mathbb{E}[L_{\mathcal{T}}^{\mu}(\mathcal{G},F)] + \beta \tau + \epsilon)-\frac{\lambda^{2}\sigma^{2}}{2} \label{eq:kl>}
\end{split}
\end{equation}

The step $c$ is because of the introduce of Lemma 1. By taking expectation over the whole hypothesis space $f_{\theta} \in \mathcal{F}$, we can get (\ref{eq:lun<lsup}). Take it into the step $b$ and the step $c$ can be drawn.
\begin{equation}
	\mathbb{E}[L_{\mathcal{P}}(\bar{\mathcal{G}},\bar{F})] \leq \gamma(1-\tau) \mathbb{E}[L_{\mathcal{T}}^{\mu}(\mathcal{G},F)] + \beta \tau + \epsilon \label{eq:lun<lsup}
\end{equation}

From the last step of Eq.(\ref{eq:kl>}), we know that the left KL divergence term is always larger than the right term for any $\lambda \in \mathbb{R}$. Taking the first derivative of the right term with respect to $\lambda$, we find that when $\lambda=\Delta / \sigma^{2}$, the maximum value is $\Delta^2 / 2 \sigma^{2}$, where $\Delta  \triangleq \mathbb{E}[L_{\mathcal{P}}(\mathcal{G},F)]-\gamma(1-\tau) \mathbb{E}[L_{\mathcal{T}}^{\mu}(\mathcal{G},F)] + \beta \tau + \epsilon$. It gives Eq.(\ref{eq:sqrt_kl_>}).

\begin{equation}
\begin{split}
& \sqrt{2\sigma^{2}D\left(P_{\mathcal{G}, F} \| P_{\mathcal{G}} \otimes P_{F}\right)} \\
& \geq 
|(\gamma(1-\tau) \mathbb{E}[ L_{\mathcal{T}}^{\mu}(\mathcal{G},F)] +\beta \tau  +\epsilon)-\mathbb{E}[L_{\mathcal{P}}(\mathcal{G},F)]| \label{eq:sqrt_kl_>}
\end{split}
\end{equation}

The relation between KL divergence and mutual information is shown in Eq.(\ref{eq:mi_kl})
\begin{equation}
I(\mathcal{G}, F) = D\left(P_{\mathcal{G}, F} \| P_{\mathcal{G}} \otimes P_{F}\right) \label{eq:mi_kl}
\end{equation}


By taking Eq.(\ref{eq:mi_kl}) into Eq.(\ref{eq:sqrt_kl_>}), if $(\gamma(1-\tau) \mathbb{E}[ L_{\mathcal{T}}^{\mu}(\mathcal{G},F)] +\beta \tau  +\epsilon)-\mathbb{E}[L_{\mathcal{P}}(\mathcal{G},F)] \geq 0$, we can get,
\begin{small}
\begin{equation}
\mathbb{E}\left[L_{\mathcal{T}}^{\mu}(\mathcal{G},F)-\frac{L_{\mathcal{P}}(\mathcal{G},F)}{\gamma(1-\tau)}\right] \leq \frac{\sqrt{2\sigma^{2}I(\mathcal{G}, F)}-(\beta \tau +\epsilon)}{\gamma(1-\tau)}
\end{equation}
\end{small}

Eventually, by taking Eq.(\ref{mc_lr}), we can meet our conclusion:
\begin{equation}
\begin{split}
\operatorname{gen}\left(\mathcal{T}, P_{F \mid \mathcal{P}}\right) & \triangleq \mathbb{E}\left[L_{\mathcal{T}}(\mathcal{G},F)-\frac{L_{\mathcal{P}}(\mathcal{G},F)}{\gamma(1-\tau)}\right]\\
\operatorname{gen}\left(\mathcal{T}, P_{F \mid \mathcal{P}}\right) &\leq \frac{\sqrt{2\sigma^{2}I(\mathcal{G}, F)}-(\beta \tau +\epsilon)}{\gamma(1-\tau)}
\end{split}
\end{equation}

\section{Proof for Contradiction} \label{Proof_for_Contradiction}

This proof follows the assumptions from GRACE \cite{DBLP:journals/corr/abs-2006-04131}. The three random variables of original graph, the embeddings of the first view and the embeddings of the second view are denoted as $\mathcal{G}$, $U$ and $V$. There exist two core Markov relations. The first is $U \leftarrow \mathcal{G} \rightarrow V$, which is Markov equivalent to $U \rightarrow \mathcal{G} \rightarrow V$ since $U$ and $V$ are conditionally independent after observing $\mathcal{G}$. The second is $\mathcal{G} \rightarrow(U, V) \rightarrow U$. Then by data processing inequality \cite{DBLP:journals/tit/Raginsky16}, The first Markov relation can lead to $I(U ; V) \leq I(U ; \mathcal{G})$, and the second  can lead to $I(\mathcal{G} ; U) \leq I(\mathcal{G} ; U, V)$. Finally, the combination of the two inequalities yields,
\begin{equation}
    I(U ; V) \leq I(\mathcal{G} ; U, V) \label{Grace_MI}
\end{equation}

The contrastive InfoNCE loss denoted as $I_{NCE}$, has been proved to be a lower bound for MI between the representation of two views \cite{DBLP:conf/icml/PooleOOAT19,agakov2004algorithm}. Then by the conclusion in Eq.(\ref{Grace_MI}), $I_{NCE}$ is also a lower bound for the MI between original graph and the representations of views $I(\mathcal{G} ; U, V)$.
\begin{equation}
    I_{NCE} \leq I(U ; V) \leq I(\mathcal{G} ; U, V)
\end{equation}

Noting that $I(\mathcal{G} ; U, V)= I(\mathcal{G} ; F(\mathcal{G}_{view}))$, we finally reach the conclusion that $I_{NCE}$ is the lower bound for MI between original graph and the GNN encoder learned hypothesis.

\section{Proof for Loss Functions and MI Bounds} \label{Proof_for_Loss_MI}
In this section, we focus on every part of our loss function defined in Eq.(\ref{eq:loss_func_f_f}) and show their relation with MI bounds. The first part is a InfoNCE term, which is a modified version of loss from GRACE \cite{DBLP:journals/corr/abs-2006-04131}. The second part is a KL term. We show that the first part of our loss function is a MI lower bound and the second part is a MI upper bound.

\subsection{Modified GRACE Loss and MI Lower Bound}
The original loss function of GRACE and our modified version are each defined below, where $\theta$ is similarity measure function, $\rho_{r}\left(\boldsymbol{u}_{i}\right)=$ $\sum_{j=1}^{N} \mathbbm{1}_{[i \neq j]} \exp \left(\theta\left(\boldsymbol{u}_{i}, \boldsymbol{u}_{j}\right) / \tau\right)$, and  $\rho_{c}\left(\boldsymbol{u}_{i}\right)=\sum_{j=1}^{N} \exp \left(\theta\left(\boldsymbol{u}_{i}, \boldsymbol{v}_{j}\right) / \tau\right)$. 

\begin{scriptsize}
\begin{equation}
\begin{split}
\mathcal{J}_{GRACE} & = \mathbb{E}_{\Pi_{i} p\left(\boldsymbol{u}_{i}, \boldsymbol{v}_{i}\right)}\left[\frac{1}{N} \sum_{i=1}^{N} \log \frac{\exp \left(\theta\left(\boldsymbol{u}_{i}, \boldsymbol{v}_{i}\right) / \tau\right)}{Base}\right]\\
Base & = \sqrt{\left(\rho_{c}\left(\boldsymbol{u}_{i}\right)+\rho_{r}\left(\boldsymbol{u}_{i}\right)\right)\left(\rho_{c}\left(\boldsymbol{v}_{i}\right)+\rho_{r}\left(\boldsymbol{v}_{i}\right)\right)}
\end{split}
\end{equation}
\end{scriptsize}

\begin{scriptsize}
\begin{equation}
\begin{split}
\mathcal{J}_{OUR} & = \mathbb{E}_{\Pi_{i} p\left(\boldsymbol{u}_{i}, \boldsymbol{v}_{i}\right)}\left[\frac{1}{N} \sum_{i=1}^{N} \log \frac{\exp \left(\theta\left(\boldsymbol{u}_{i}, \boldsymbol{v}_{i}\right) / \tau\right)}{Base^{\prime}}\right]\\
Base^{\prime} &= \sqrt{\left(\rho_{c}\left(\boldsymbol{u}_{i}+\epsilon \right)+\rho_{r}\left(\boldsymbol{u}_{i}\right)\right)\left(\rho_{c}\left(\boldsymbol{v}_{i} +\epsilon \right)+\rho_{r}\left(\boldsymbol{v}_{i}\right)\right)}
\end{split}
\end{equation}
\end{scriptsize}

We take inner product as $\theta$, by the definition of $\rho_{c}$ we have,
\begin{equation}
\begin{split}
\rho_{c}\left(\boldsymbol{u}_{i} +\epsilon \right)&=\sum_{j=1}^{N} \exp \left(\theta\left(\boldsymbol{u}_{i} +\epsilon , \boldsymbol{v}_{j}\right) / \tau\right)\\ 
&=\sum_{j=1}^{N} \exp \left( \langle \boldsymbol{u}_{i} +\epsilon , \boldsymbol{v}_{j} \rangle / \tau\right)\\
&=\sum_{j=1}^{N} \exp \left( \langle \boldsymbol{u}_{i}, \boldsymbol{v}_{j} \rangle / \tau\right) + \exp \left( \langle \epsilon , \boldsymbol{v}_{j} \rangle / \tau\right)
\end{split}
\end{equation}

The second term of the last step is $\mathbb{E}[\epsilon V]$, the expectation of the similarity between the standard normal distribution noise and the learned embeddings. Moreover, $\mathbb{E}[\epsilon V]=\mathbb{E}\left[\epsilon\right] \cdot \mathbb{E}[V]=0$ since the embeddings are independent with the noise. Therefore, the noise term similarity can be ignored in $\rho_{c}\left(\boldsymbol{u}_{i} +\epsilon \right)$, which also holds for $\rho_{c}\left(\boldsymbol{v}_{i} +\epsilon \right)$.  Eventually we have $\mathcal{J}_{GRACE} \approx   \mathcal{J}_{OUR}$, and by the proof from GRACE, our modified loss is also a lower bound of MI between two view representations.

\subsection{KL term and MI Upper Bound}
The KL loss has been proved to be a upper bound of MI  \cite{DBLP:conf/icml/PooleOOAT19,agakov2004algorithm}, which indicates that, if $X$ and $Y$ are denoted as two random variables, then the inequality $ I(X ; Y) \leq \mathbb{E}_{p(x)}[K L(p(y \mid x) \mid q(y))] $ holds. By setting $X = \mathcal{G}_{view}$, $\; Y = F(\mathcal{G}_{view})$  and  $q(y)$ be the standard normal distribution $\mathcal{N}(0,1)$, we can meet the conclusion that the KL regularization term is the upper bound of the MI between graph views and the GNN encoder learned hypothesis.

\section{Algorithmic Format for InfoAdv} \label{algor}

Algorithm \ref{InfoAdv_Algorithm} describes the algorithmic format for InfoAdv. InfoAdv is a self-supervised training algorithm with view generator $t_{\phi}$ and target encoder $f_{\theta}$. 

\SetKwInOut{KwResult}{HyperParams}
\begin{algorithm*}[h]  
	\caption{Training Generalizable GCL with InfoAdv Framework}
	\label{InfoAdv_Algorithm}
	\LinesNumbered 
	\KwIn{Graph feature matrix $\boldsymbol{X}$; Adjacency matrix $\boldsymbol{A}$; Target encoder $f_{\theta}$; view generator $t_{\phi}$}
	\KwResult{Feature-mask probability $(p_{f_1},p_{f_2})$; Edge-drop probability $(p_{ea},p_{e_2})$; Loss balance $\lambda$; Learning rates $\alpha,\beta$ }
	\KwOut{Trained target encoder $f_{\theta^{*}}$}
	
	\tcc{define losses for single node}

    \textbf{define} $\ell_{1}\left(\boldsymbol{u}_{i}^{\mu}, \boldsymbol{v}_{i}^{\mu}, \boldsymbol{u}_{i}\right)=-\log \frac{e^{\theta\left(\boldsymbol{u}_{i}^{\mu}, \boldsymbol{v}_{i}^{\mu} \right) / \tau}}{e^{\theta\left(\boldsymbol{u}_{i}^{\mu}, \boldsymbol{v}_{i}^{\mu} \right) / \tau}+\sum_{k=1}^{N} \mathbbm{1}_{[k \neq i]} e^{\theta\left(\boldsymbol{u}_{i}, \boldsymbol{v}_{k}^{\mu} \right) / \tau}+\sum_{k=1}^{N} \mathbbm{1}_{[k \neq i]} e^{\theta\left(\boldsymbol{u}_{i}^{\mu}, \boldsymbol{u}_{k}^{\mu}\right) / \tau}}$\;

    \textbf{define} $\ell_{2}(\boldsymbol{u}_{i}^{\sigma},\boldsymbol{u}_{i}^{\mu}) = \frac{1}{2}\left[{(\boldsymbol{u}_{i}^{\sigma})}^{2}+{(\boldsymbol{u}_{i}^{\mu})}^{2}-2\log \left( \boldsymbol{u}_{i}^{\sigma} \right)-1\right]$

    \tcc{calculate loss for every node}
    
    $\mathcal{J}_1=\frac{1}{2 N} \sum_{i=1}^{N}\left[\ell_{1}\left(\boldsymbol{u}_{i}^{\mu}, \boldsymbol{v}_{i}^{\mu}, \boldsymbol{u}_{i} \right)+\ell_{1}\left(\boldsymbol{v}_{i}^{\mu}, \boldsymbol{u}_{i}^{\mu}, \boldsymbol{v}_{i}\right)\right]$
    
    $\mathcal{J}_2=\frac{1}{2 N} \sum_{i=1}^{N} \left[\ell_{2}(\boldsymbol{u}^{\mu}_{i},\boldsymbol{u}^{\sigma}_{i})+\ell_{2}(\boldsymbol{v}^{\mu}_{i},\boldsymbol{v}^{\sigma}_{i})\right]$\;
    
    $\mathcal{J}_2^{\prime}=\frac{1}{N} \sum_{i=1}^{N} \ell_{2}(\boldsymbol{u}^{\mu}_{i},\boldsymbol{u}^{\sigma}_{i})$\;
    
    $\mathcal{J}=\mathcal{J}_1+ \lambda \mathcal{J}_2$\;
		
	\For{epoch $\leftarrow 1,2,\dots$ }{
		\tcc{obtain the learnable view via view generator}
		$\widetilde{\boldsymbol{A}}_{1} \leftarrow t_{\phi}(\boldsymbol{X},\boldsymbol{A},p_{ea});$
		
		$\widetilde{\boldsymbol{X}}_{1} \leftarrow dropout(\boldsymbol{X},p_{f_1});$
		
		\tcc{obtain the random view via random dropout}
		
		$\widetilde{\boldsymbol{A}}_{2} \leftarrow dropout(\boldsymbol{A},p_{e_2});$
		
		$\widetilde{\boldsymbol{X}}_{2} \leftarrow dropout(\boldsymbol{X},p_{f_2});$ 
		
		\tcc{obtain the embeddings of two views}
		
		$\boldsymbol{U},\boldsymbol{U}^{\mu},\boldsymbol{U}^{\sigma} \leftarrow  f_{\theta}(\widetilde{\boldsymbol{X}}_{1},\widetilde{\boldsymbol{A}}_{1});$
		
		$\boldsymbol{V},\boldsymbol{V}^{\mu},\boldsymbol{V}^{\sigma}  \leftarrow  f_{\theta}(\widetilde{\boldsymbol{X}}_{2},\widetilde{\boldsymbol{A}}_{2});$
		
		
		\tcc{update augmenter parameters via gradient ascent}
		
		$\phi \leftarrow \phi-\alpha \nabla_{\phi} (\mathcal{J}_2^{\prime});$
		
		\tcc{update target encoder \& projection head  parameters via gradient descent}
		
		$\theta \leftarrow \theta-\beta \nabla_{\theta}(\mathcal{J});$
		
		$\omega \leftarrow \omega-\beta \nabla_{\omega}(\mathcal{J});$
	}
	\Return{Target encoder $f_{\theta}$}
\end{algorithm*}

\section{Datasets Description} \label{datasets}

Download links are provided in Table \ref{Dataset_download_links}.
The statistics of datasets used in experiments are summarized in Table \ref{Dataset_statistics}.  We use two kinds of eight widely-used datasets, they are: (1) Citation networks including Cora \cite{DBLP:journals/ir/McCallumNRS00}, Citeseer \cite{DBLP:conf/dl/GilesBL98}, Pubmed \cite{DBLP:journals/aim/SenNBGGE08}, Coauthor-CS \cite{DBLP:journals/corr/abs-1811-05868}, Coauthor-Phy \cite{DBLP:journals/corr/abs-1811-05868}. (2) Social networks including WikiCS \cite{DBLP:journals/corr/abs-2007-02901}, Amazon-Photos \cite{DBLP:journals/corr/abs-1811-05868}, Amazon-Computers \cite{DBLP:journals/corr/abs-1811-05868}.

\begin{table}[!htbp]
  \caption{Dataset statistics}
  \label{Dataset_statistics}
  \centering
  \setlength{\tabcolsep}{1mm}{
\begin{tabular}{ccccc}
\toprule 
Dataset & \#Nodes & \#Edges & \#Features & \#Classes \\
\midrule
Cora  & 2,708 & 5,429 & 1,433 & 7 \\
Citeseer  & 3,327 & 4,732 & 3,703 & 6\\
Pubmed  & 19,717 & 44,338 & 500 & 3\\
Coauthor-CS  & 18,333 & 81,894 & 6,805 & 15 \\
Coauthor-Physics  & 34,493 & 247,962 & 8,415 & 5 \\
\midrule
WikiCS  & 11,701 & 216,123 & 300 & 10 \\
Amazon-Photo  & 7,650 & 119,081 & 745 & 8 \\
Amazon-Computers  & 13,752 & 245,861 & 767 & 10 \\
\bottomrule
\end{tabular}}
\end{table}

\begin{table*}[!htbp]
  \caption{Dataset download links}
  \centering
  \label{Dataset_download_links}
  \small
  \setlength{\tabcolsep}{3mm}{
\begin{tabular}{cl}
\toprule 
Dataset & Download Link \\
\midrule 
Cora & \href{https://github.com/kimiyoung/planetoid/raw/master/data}{https://github.com/kimiyoung/planetoid/raw/master/data} \\
Citeseer & \href{https://github.com//kimiyoung/planetoid/raw/master/data}{https://github.com//kimiyoung/planetoid/raw/master/data} \\
Pubmed & \href{https://github.com/kimiyoung/planetoid/raw/master/data}{https://github.com/kimiyoung/planetoid/raw/master/data} \\
Coauthor-CS & \href{https://github.com/shchur/gnn-benchmark/raw/master/data/npz/ms\_academic\_cs.npz}{https://github.com/shchur/gnn-benchmark/raw/master/data/npz/ms\_academic\_cs.npz} \\
Coauthor-Phy &  \href{https://github.com/shchur/gnn-benchmark/raw/master/data/npz/ms\_academic\_phy.npz}{https://github.com/shchur/gnn-benchmark/raw/master/data/npz/ms\_academic\_phy.npz} \\
\midrule 
WikiCS & \href{https://github.com/pmernyei/wiki-cs-dataset/raw/master/dataset}{https://github.com/pmernyei/wiki-cs-dataset/raw/master/dataset} \\
Amazon-Photo & \href{https://github.com/shchur/gnn-benchmark/raw/master/data\\/npz/amazon\_electronics\_photo.npz}{https://github.com/shchur/gnn-benchmark/raw/master/data/npz/amazon\_electronics\_photo.npz}\\
Amazon-Computers & \href{https://github.com/shchur/gnn-benchmark/raw/master/data/npz/amazon\_electronics\_computers.npz}{https://github.com/shchur/gnn-benchmark/raw/master/data/npz/amazon\_electronics\_computers.npz} \\
\bottomrule
\end{tabular}}
\end{table*}

\begin{table*}[h]
  \caption{Hyperparameters settings}
  \centering
  \small
  \label{Hyperparameters_Table}
  \setlength{\tabcolsep}{1.8mm}{
\begin{tabular}{cccccccccc}
\toprule 
Dataset & $\lambda$ & $(p_{f_1},p_{f_2})$ & $(p_{ea},p_{e_2})$ & Learning rate & Weight decay & Epochs & Hidden dim & Activation \\
\midrule 
Cora & $10^{-5}$ & (0.4,0.3) & (0.8,0.2) & $0.0005$ & $10^{-5}$ & 1000 & 128 & ReLU \\
Citeseer & $10^{-5}$ &(0.3,0.2) & (0.2,0.0) & $0.001$ & $10^{-5}$ & 500 & 256 & PReLU \\
Pubmed & 10 & (0.1,0.1) & (0.3,0.5) & $0.001$ & $10^{-5}$ & 2500 & 256 & ReLU \\
Coauthor-CS & $10^{-5}$ & (0.3,0.4) & (0.3,0.2) & $0.0005$ & $10^{-5}$ & 1000 & 256 & RReLU \\
Coauthor-Phy & 10 & (0.1,0.4) & (0.4,0.1) & $0.01$ & $10^{-5}$ & 1900 & 128 & RReLU \\
\midrule 
Wiki-CS & 10 & (0.1,0.1) & (0.2,0.3) & $0.01$ & $10^{-5}$ & 3100 & 256 & PReLU \\
Amazon-Photo & 60 & (0.1,0.1) & (0.9,0.3) & $0.01$ & $10^{-5}$& 2700 & 256 & ReLU \\
Amazon-Computers & 0.5 & (0.2,0.3) & (0.9,0.3)&  $0.01$ & $10^{-5}$ & 2000 & 128 & RReLU \\
\bottomrule
\end{tabular}}
\end{table*}

\begin{table*}[h]
  \caption{Noise hyperparameter, graph density and noise rate}
    \centering
  \small
  \label{Noise_Rate_Table}
  \setlength{\tabcolsep}{0.5mm}{
\begin{tabular}{ccccccccccccc}
\toprule 
\multirow{3}*{Cora} & Hyperparam & 0.0 & 0.0010 & 0.0011 & 0.0012 & 0.0013 & 0.0014 & 0.0015 & 0.0016 & 0.0017 & 0.0018 & 0.0019 \\
~ & Density & 0.14\% & 0.24\% & 0.25\% & 0.26\%& 0.27\%& 0.28\%& 0.29\%& 0.30\%& 0.31\%& 0.32\% & 0.33\% \\
~ & Noise rate & +0\% & +71.4\% & 78.5\% & +85.7\% & +92.8\% & +100\% & +107.1\% & +114.2\% & +121.4\% & +128.5\% & +135.7\% \\
\midrule
\multirow{3}*{Amazon-Photo} & Hyperparam & 0.0 & 0.0010 & 0.0015 & 0.0020 & 0.0025 & 0.0030 & 0.0035 & 0.0040 & 0.0045 & 0.0050 & 0.0055 \\
~ & Density & 0.41\% & 0.51\% & 0.56\% & 0.61\% & 0.66\% & 0.71\% & 0.76\% & 0.80\% & 0.85\% & 0.90\% & 0.95\% \\
~ & Noise rate & +0\% & +24.3\% & +36.5\% & +48.7\% & +60.9\% & +73.3\% & +85.6\% & +97.8\% & +110.0\% & +122.3\% & +134.5\% \\
\bottomrule
\end{tabular}}
\end{table*}

\section{Experiments Setting} \label{setting}

\subsection{Evaluation Protocol} \label{Eva_Subsection}

\subsubsection{Node classification}
For task of node classification, our evaluation protocol is the same as \citet{DBLP:journals/corr/abs-2006-04131,DBLP:conf/www/0001XYLWW21}, which follows the linear evaluation scheme as introduced in \citet{DBLP:conf/iclr/VelickovicFHLBH19}, where each model is firstly trained in an unsupervised manner, then, the resulting representations are used to train and test a simple $l_2$-regularized logistic regression classifier. We randomly select 10\% of the nodes as the training set, where a 5-fold cross-validation is conducted and leave the rest nodes as the testing set. The performance metric used in our experiments is the $F_1$-micro score. Moreover, for fair evaluation, we train each model for three different data splits for every datasets and report its average performance and standard deviation. 

\subsubsection{Link predication}
For task of link predication, we set the original edges in the graph as positive samples and split them into training set, validation set and testing set in the ratio of 85\%, 5\%, and 10\%. The node representations is learned only with training set edges. During validation and testing, negative samples is randomly selected from originally unconnected node pairs. The positive-negative sample ratio is kept as 1:1. We calculate the link probability via the inner product of the representations and adopt Area Under Curve(AUC) and Average Precision(AP) as the evaluation metrics.

\subsection{Baselines} \label{Baseline_Subsection}

We consider representative baseline methods belonging to the following two categories, self-supervised learning and supervised learning. Self-supervised learning methods include,
\begin{itemize}
	\item [(1)] Traditional graph embedding methods include DeepWalk \cite{DBLP:conf/kdd/PerozziAS14}, Raw-features and the concatenation of DeepWalk with Raw-features.
	\item [(2)] Generative(Predictive) learning methods, such as Graph Auto-encoders (GAE,VGAE) \cite{DBLP:journals/corr/KipfW16a}.
	\item [(3)] Contrastive learning methods include Deep Graph InfoMax (DGI) \cite{DBLP:conf/iclr/VelickovicFHLBH19}, GRACE \cite{DBLP:journals/corr/abs-2006-04131}, GCA \cite{DBLP:conf/www/0001XYLWW21}, and AD-GCL \cite{DBLP:conf/nips/SureshLHN21}. Noting that, the original AD-GCL only focus on graph-level classification task, and we modify AD-GCL to node-level classification to conduct comparative experiments.
\end{itemize}
Supervised learning methods include Graph Convolutional Networks (GCN)\cite{DBLP:conf/iclr/KipfW17} and Graph Attention Networks (GAT)\cite{DBLP:conf/iclr/VelickovicCCRLB18}, where they are trained transductively in an end-to-end fashion.


\subsection{Hyperparameters} \label{Hyper_Subsection}

The hyperparameter settings is shown in Table  \ref{Hyperparameters_Table}. The key hyperparameters are the feature-mask probability $(p_{f_1},p_{f_2})$, the edge-drop probability $(p_{ea},p_{e_2})$ and the loss balance $\lambda$, which are selected by multiple experiments. The synchronously increasing relation of noise hyperparameter, graph density and the noise rate is shown in Table \ref{Noise_Rate_Table}. The noise hyperparameter is a probability threshold of introducing noise. The graph density is the non-zero rate of graph adjacency matrix. The noise rate is the ratio of noise edge number and original edge number.

\subsection{Computing Resources}  \label{Compres_Subsection}
All our experiments are performed on RedHat server (4.8.5-39) with Intel(R) Xeon(R) Gold 5218 CPU $@$ 2.30GHz4 and $4 \times$ NVIDIA Tesla V100 SXM2 (32GB)

\section{Limitations and Future Explorations} \label{Limit_Section}
The fundamental limitation of our work is that, it is under the paradigm of self-supervised learning, which has no information from downstream tasks. Therefore, it is hard for us to get rid of all downstream irrelevant information during the GCL pretext task fitting and reach the best generalization. However, our GCL-GE metric and its MI upper bound do provide a novel guideline principle for reaching better GCL generalization in an information-theoretic spirit. Moreover, our proposed InfoAdv framework does achieve this goal from two GCL sub-parts, the learnable augmentation and the contrastive loss.

The second limitation of our work is the paradigm restriction. The restriction comes from the proof of Lemma 1, which introduces a latent class assumption from \citet{DBLP:conf/icml/SaunshiPAKK19}. This assumption is built on positive/negative samples and thus heavily depends on the traditional contrastive learning paradigm. However, the GCL-GE metric itself is not bound with contrastive learning, which leaves us a future exploration to extent our principle to other learning paradigm like predictive(generative) learning. To achieve this goal, a new theoretical explanation is in need to unify the predictive(generative) learning and contrastive learning.

\begin{figure*}[!htbp]
  \centering
  \includegraphics[width=0.9\linewidth]{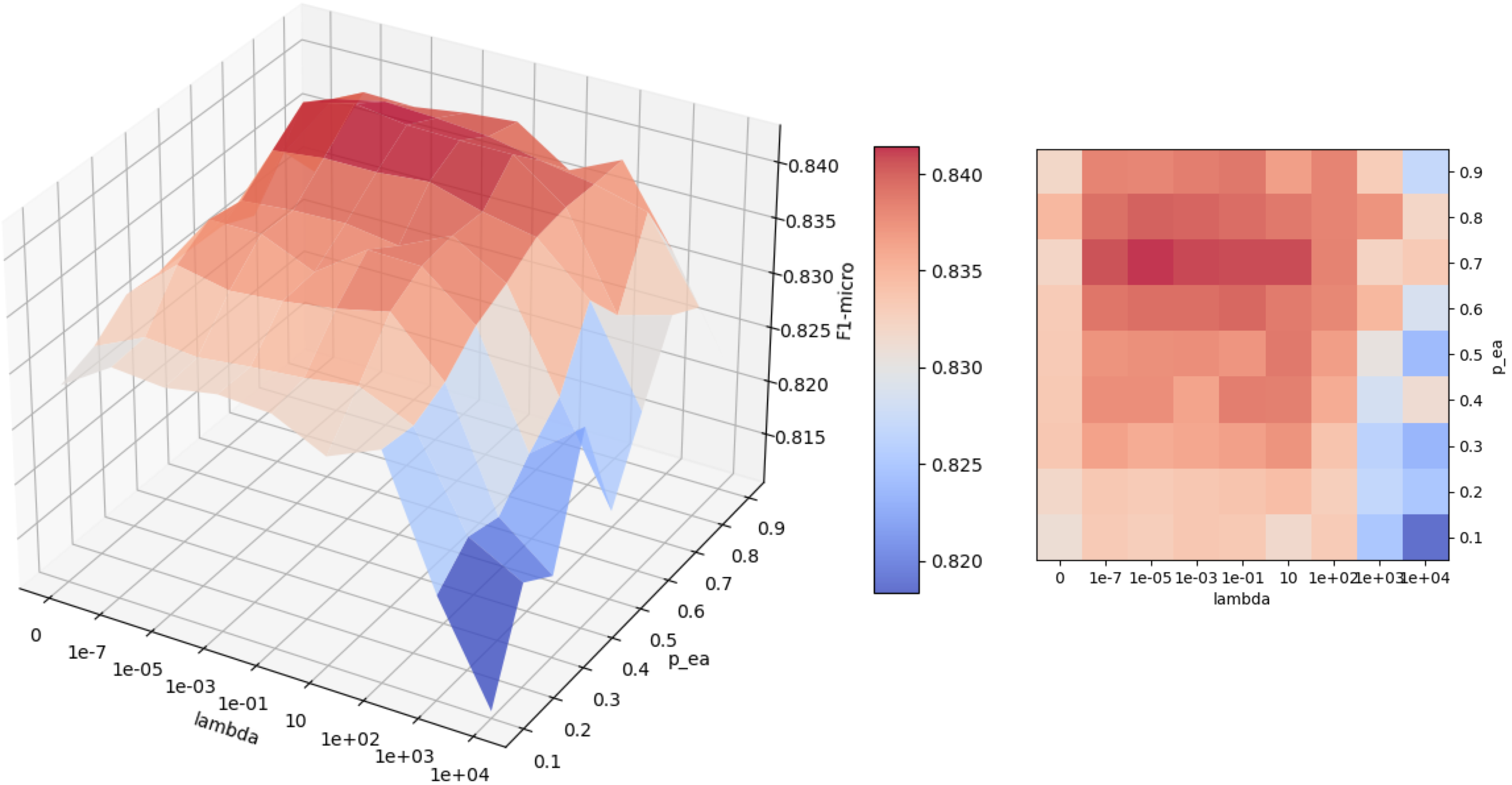}
  \captionsetup{font=small}
  \caption{The \textbf{left} side is the 3D-surface for the performance on Cora dataset, whose three axis are the balance for KL regularization term $\lambda$, the edge-drop probability of the view learned from view generator $p_{ea}$, and the performance $F_1$-micro score. The \textbf{middle} is the colorbar for the performance, indicating that the warmer the color is, the better the model performs. The \textbf{right} side is the heatmap for the 3D-surface when it is observed from the top, i.e., $F_1$-micro axis.}   \label{hyper_lamb2}
\end{figure*}

\begin{figure*}[!htbp]
  \centering
  \includegraphics[width=0.9\linewidth]{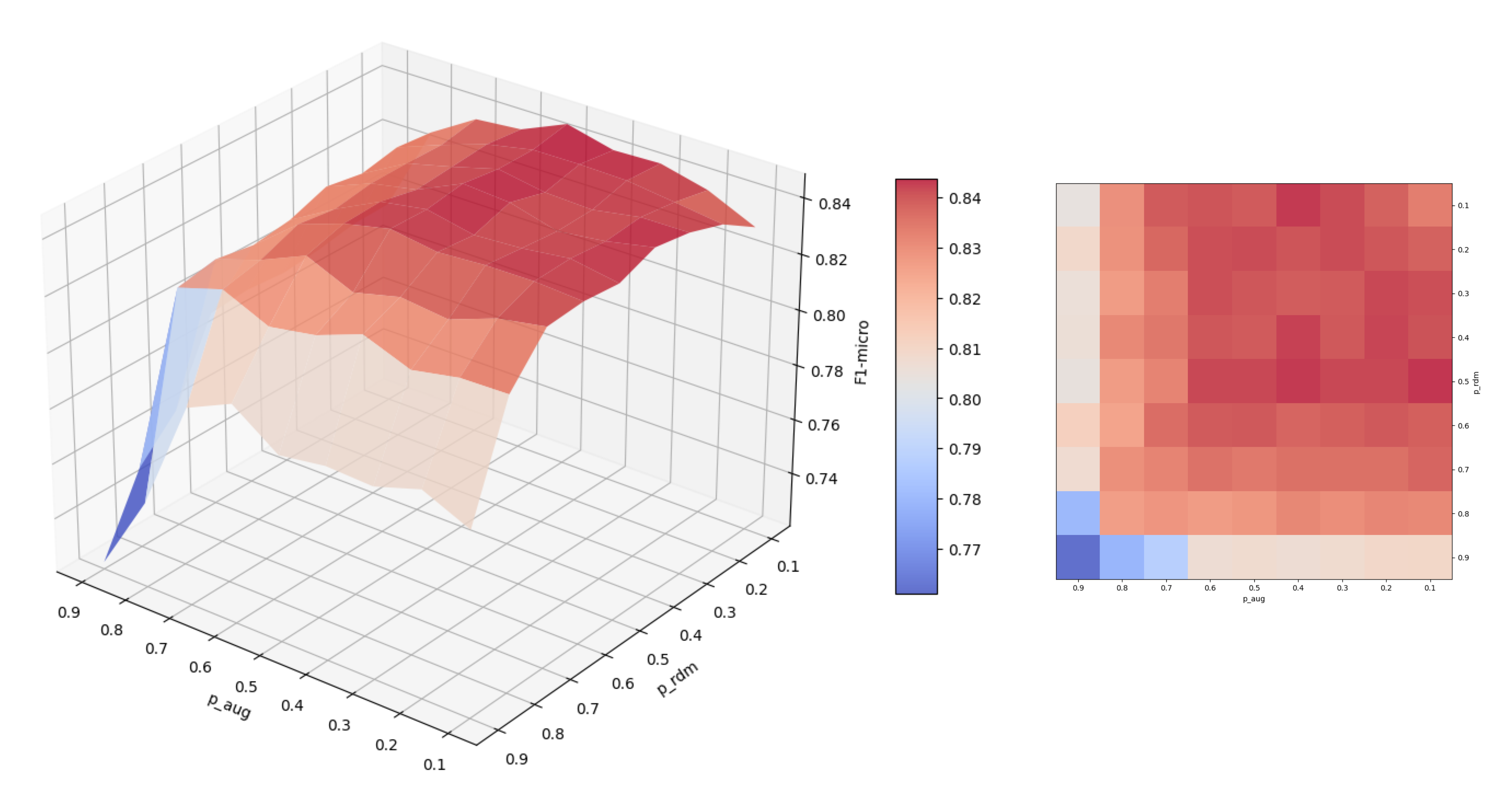}
  \captionsetup{font=small}
  \caption{The \textbf{left} side is the 3D-surface for the performance on Cora dateset, whose three axis are the probability of the learned view $p_{aug}$, the probability of the random view $p_{rdm}$, and the performance $F_1$-micro score. The \textbf{middle} is the colorbar for the performance, indicating that the warmer the color is, the better the model performs. The \textbf{right} side is the heatmap for the 3D-surface when it is observed from the top, i.e., $F_1$-micro axis.} \label{hyper_rates}
\end{figure*}

\section{Additional Experiments for Hyperparameter Sensitivity} \label{Sensitivity_Section}

This section is devoted to sensitivity analysis on hyperparameters. The hyperparameters consist of the feature-mask probability of two views $(p_{f_1},p_{f_2})$, the edge-drop probability of two views $(p_{ea},p_{e_2})$ and the balance for KL regularization term $\lambda$. Among them, the two most critical hyperparameters are, (1) $p_{ea}$ the edge-drop probability of the view learned from view generator. (2) $\lambda$ the balance for KL regularization term.

\subsection{Critical Hyperparameters Sensitivity}

The target hyperparameters of this experiment is the two critical hyperparameters $\lambda$ and $p_{ea}$.
We show the model sensitivity on the transductive node classification task on Cora, under the perturbation of these two hyperparameters, that is, varying $\lambda$ from $1e-07$ to $1e+04$ and $p_{ea}$ from 0.1 to 0.9 to plot the 3D-surface for performance in terms of $F_1$-micro.

The result is shown in Figure \ref{hyper_lamb} (also Figure \ref{hyper_lamb2}), where four key conclusions can be observed,
\begin{itemize}
    \item [(1)] The model is relatively stable when $\lambda$ is not too large, as the model performance remains steady when $\lambda$ varies in the range of $1e-07$ to $1e+02$.
    \item [(2)] The model stays stable even when $\lambda$ is very small, since the model maintains a high level of performance even $\lambda$ reachs as small as $1e-07$.
    \item [(3)] Higher edge-drop probability of the view learned from view generator helps model reach better performance, as the best performance occurs when $p_{ea}=0.7$.
    \item [(4)] The KL regularization component does work. If we do not use the KL term, that is to set $\lambda=0$, the performance drops sharply.
\end{itemize}

\subsection{View Probabilities Sensitivity}
The target hyperparameters of this experiment is the four probabilities $(p_{f_1},p_{f_2})$ and $(p_{ea},p_{e_2})$.
We show the model sensitivity on the transductive node classification task on Cora, under the perturbation of these four hyperparameters. Specifically, we set $p_{aug}=p_{f_1}=p_{ea}$ and $p_{rdm}=p_{f_2}=p_{e_2}$ for sake of visualization brevity \cite{DBLP:journals/corr/abs-2006-04131}, and vary $p_1$ and $p_2$ from 0.1 to 0.9 to plot the 3D-surface for performance in terms of $F_1$-micro.

The result is shown in Figure \ref{hyper_rates},from which we can learn two things,
\begin{itemize}
    \item [(1)] The model is relatively stable when the view probabilities are not too large, since it can be observed that the model performance stays on the plateau in most cases, and only drops sharply when all probabilities are large than 0.8.
    \item [(2)] The best performance occurs neither in the smallest probabilities nor the largest probabilities. It is reached in some middle position.
\end{itemize}

\begin{figure*}
  \centering
  \includegraphics[width=1\linewidth]{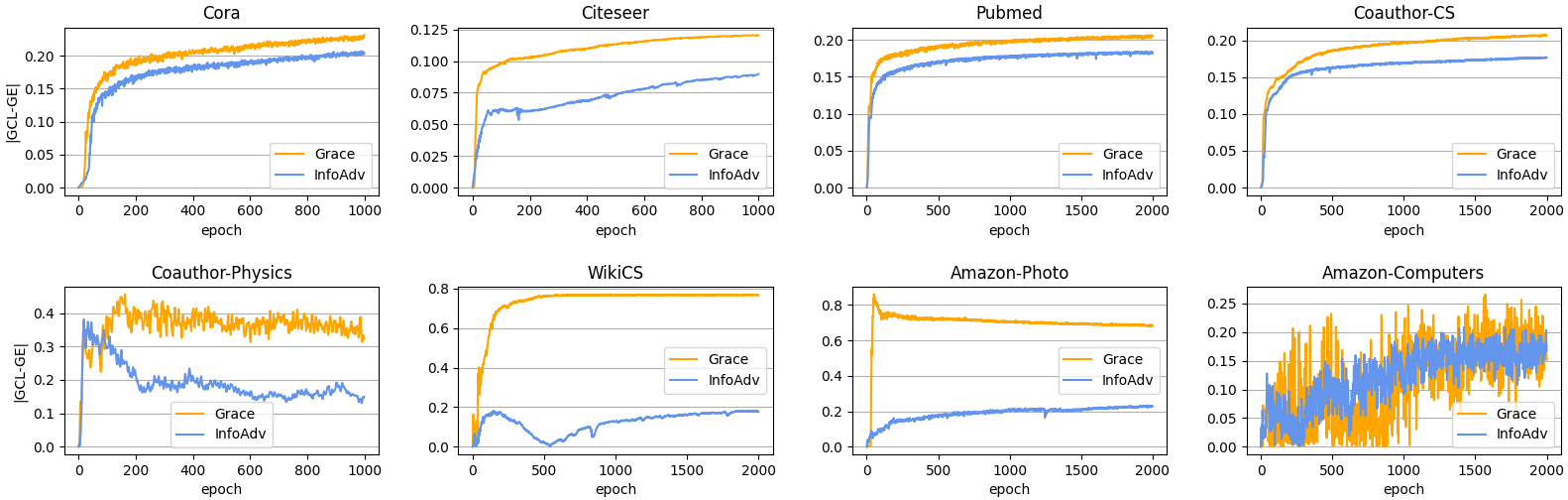}
  \captionsetup{font=small}
  \caption{The absolute value of GCL-GE of InfoAdv and baseline during the whole training procedure on eight datasets. The x-axis is the training epoch. The y-axis is the absolute value of GCL-GE.} \label{GCLGE_metric}
\end{figure*}

\section{Additional Experiments for GCL-GE Metric} \label{Metric_Section}

This section is devoted to analysis on our proposed GCL-GE metric. In this experiment, we compare the GCL-GE empirical value between InfoAdv and baseline GRACE. The value is calculated according to the definition in Eq.(\ref{eq:ge}), and is achieved by following specific implementation. (1) Both the pretext task loss and the downstream task loss are normalized by a normalization term. The normalization term is an empirical value of maximum loss, which can be approximately calculated by the loss of model with random parameter, E.g., the first epoch. This normalization is a realization of $k$ in Eq.(\ref{eq:ge}), rescaling the two losses into the same scale and enabling the subtraction. (2) The downstream task is not fitted by logistic regression classifier. Instead, we use Mean Classifier defined in Appendix \ref{2diff}.

The result is shown in Figure \ref{GCLGE_metric}, from which it can be observed that the absolute value of GCL-GE of InfoAdv is lower than baseline GRACE on all eight datasets. These phenomena reveal that our InfoAdv performs better than baseline in terms of GCL-GE metric, which demonstrates that our InfoAdv dose improve generalization ability.

\end{document}